\documentclass[sigconf]{acmart}

\usepackage{graphicx}
\usepackage{amsfonts}            %数学字体
\usepackage{algorithm}
\usepackage{algorithmic}
\usepackage{multirow}
\usepackage{subfigure}
\usepackage{bm}
\usepackage{hyperref}

\usepackage{color}
\usepackage{xcolor}
\usepackage{colortbl}
\usepackage{makecell}
\AtBeginDocument{%
	\providecommand\BibTeX{{%
			\normalfont B\kern-0.5em{\scshape i\kern-0.25em b}\kern-0.8em\TeX}}}

\setcopyright{acmcopyright}
\copyrightyear{2022}
\acmYear{2022}
\setcopyright{acmcopyright}\acmConference[SIGMOD '23]{ACM SIGMOD/PODS International Conference on Management of Data}{June 18-23, 2023}{Seattle}
\acmConference[SIGMOD '23]{SIGMOD '23: ACM SIGMOD/PODS International Conference on Management of Data}{June 18-23, 2023}{SIGMOD, Seattle}
\acmBooktitle{SIGMOD '23: ACM SIGMOD/PODS International Conference on Management of Data, 
	June 18-23, 2023, Seattle, USA}
\acmPrice{15.00}
\acmISBN{978-1-4503-XXXX-X/18/06}

\begin{document}
	\fancyhead{}
	\title{Unsupervised Hashing with Semantic Concept Mining}
	
	
	\author{Rong-Cheng Tu$^{123}$, Xian-Ling Mao$^{123}$, Kevin Qinghong Lin$^{4}$, Chengfei Cai$^{5}$, Weize Qin$^{6}$, Hongfa Wang$^{6}$, Wei Wei$^{7}$, Heyan Huang$^{123}$}
	\affiliation{%
		\institution{$^1$School of Computer Science and Technology, Beijing Institute of Technology, Beijing, China}
		\institution{$^2$Beijing Engineering Research Center of High Volume Language Information Processing and Cloud Computing Applications, Beijing, China}
	\institution{$^3$Beijing Institute of Technology Southeast Academy of Information Technology, Fujian, China}
	\institution{$^4$National University of Singapore, Singapore}
	\institution{$^5$School of Computer Science, Zhejiang University, Zhejiang, China}
	\institution{$^6$Institute of Computing Technology, Chinese Academy of Sciences, Beijing, China}
	\institution{$^7$School of Computer Science, Huazhong University of Science and Technology, Wuhan, China}
	\country{ }
}
	
\email{{tu\_rc, maoxl}@bit.edu.cn,  linqinghong@email.szu.edu.cn,  happyccfnew@163.com,  }
\email{qinweize@ict.ac.cn,   hongfawang@gmail.com, weiw@hust.edu.cn,   hhy63@bit.edu.cn}
	
	\renewcommand{\shortauthors}{}
	
	\begin{abstract}
		Recently, to improve the unsupervised image retrieval performance, plenty of unsupervised hashing methods have been proposed by designing a semantic similarity matrix, which is based on the similarities between image features extracted by a pre-trained CNN model. However, most of these methods tend to ignore high-level abstract semantic concepts contained in images. Intuitively, concepts play an important role in calculating the similarity among images. In real-world scenarios, each image is associated with some concepts, and the similarity between two images will be larger if they share more identical concepts. Inspired by the above intuition, in this work, we propose a novel Unsupervised Hashing with Semantic Concept Mining, called UHSCM, which leverages a VLP model to construct a high-quality similarity matrix. Specifically, a set of randomly chosen concepts is first collected. Then, by employing a vision-language pretraining (VLP) model with the prompt engineering which has shown strong power in visual representation learning, the set of concepts is denoised according to the training images. Next, the proposed method UHSCM applies the VLP model with prompting again to mine the concept distribution of each image and construct a high-quality semantic similarity matrix based on the mined concept distributions. Finally, with the semantic similarity matrix as guiding information, a novel hashing loss with a modified contrastive loss based regularization item is proposed to optimize the hashing network. Extensive experiments on three benchmark datasets show that the proposed method outperforms the state-of-the-art baselines in the image retrieval task. The source codes are available \footnote{https://github.com/rongchengtu1/UHSCM}.
	\end{abstract}
	
	
	
	
	
	\maketitle
	
	\section{Introduction}
	Recently, with tremendous amounts of image data being generated, image retrieval techniques with fast retrieval speed have attracted much more attention. Among the existing image retrieval techniques, the hashing methods \cite{liu2012compact,liu2016multimedia,li2018self,zhang2016discrete,he2010scalable,yang2018shared,yang2017pairwise,tu2021hashing} are advantageous due to their high retrieval efficiency and low storage cost. The core idea of image hashing is to map images into compact hash codes while preserving original semantic similarity.
	
	According to whether the training data contains labelling information, existing hashing methods can be roughly divided into two categories: supervised and unsupervised hashing. The supervised ones \cite{tu2021weighted,shen2015supervised,liu2012supervised,wang2017supervised,eghbali2019deep,Luo2019DiscreteHW,wang2013learning} use the label information of the images to supervise the training of hashing models. Therefore, such methods usually achieve high retrieval performance. Nevertheless, it is time-consuming and expensive to annotate the images. Hence, recently, more and more researchers pay attention to the unsupervised hashing methods \cite{lin2021dsah,Zhan2016UnsupervisedDH,huang2017unsupervised,tu2018object,shen2020auto,tu2020mls3rduh} which train the hashing models with unlabeled images. 
	
	As there is no label information in the unsupervised setting, to improve the retrieval performance, one of the key points is to define a kind of guiding information which sufficiently contains the original similarity of images. Hence, many unsupervised hashing methods focus on defining similarity matrices as guiding information based on the image features extracted usually by a pre-trained CNN model, such as VGG19 \cite{simonyan2014very} or Alexnet \cite{krizhevsky2012imagenet}. For example, Semantic Structure-based unsupervised Deep Hashing (SSDH) \cite{yang2018semantic} utilizes the cosine similarity distribution of pairs based on the Gaussian estimation to construct the similarity structure. Furthermore, MLS$^3$RDUH \cite{tu2020mls3rduh} constructs the similarity matrix by utilizing manifold and cosine similarity between image features to reconstruct the local similarity structure.
	
	However, most of these methods define the similarity matrix only depending on the cosine similarity between image features but ignore the high-level abstract semantic concepts contained in the images. Intuitively, the concepts play an important role in calculating the similarity between images. In real-world scenarios, each image is associated with some concepts, and the similarity between two images will be larger if they share more identical concepts. Hence, the semantic concept information is useful to define the similarities between images to further improve the quality of semantic similarity matrix. Moreover, how can the concepts contained in the images be mined? Fortunately, we can leverage vision-language pre-training (VLP) models to mine the concept information contained in the images. By exploiting contrastive learning, these VLP models are directly pre-trained with large-scale noisy image-text pairs which are easily collected from the Internet. With such a broader and cheaper source of data while benefiting from the semantic lever supervision from texts, the VLP models are trained to align the semantic relationships between the images and their corresponding texts. Therefore, by transforming the concepts into texts through a prompt template, the VLP models can be used to align the semantic relationship between images and concept based texts, and then we can obtain the concept information contained in the images.
	
	Therefore, inspired by the above intuition, we propose a novel Unsupervised Hashing with Semantic Concept Mining (UHSCM), which leverages a VLP model CLIP \cite{radford2021learning} to construct a high-quality similarity matrix through the prompt engineering. Specifically, UHSCM first randomly collects a set of concepts, such as 'cat', 'dog', and 'flower'. Then, by adopting the prompt engineering,  UHSCM leverages the CLIP model to denoise the set of randomly collected concepts according to training images. Next, based on the denoised concepts, we utilize the CLIP model with the prompt engineering again to mine the concept distributions of images to construct a high-quality semantic similarity matrix. Finally,  incorporating with the semantic similarity matrix as guiding information, a novel hashing loss function with a modified contrastive loss based regularization item is proposed to optimize the hashing network to generate distinguished hash codes. 
	Above all, the main contributions of UHSCM are summarized as follows:
	\begin{itemize}
		\item To the best of our knowledge, the proposed UHSCM is the first work in deep unsupervised hashing, which utilizes a VLP model to mine the concept distributions of images to construct a high-quality semantic similarity matrix.
		\item A novel contrastive loss based regularization item is proposed for hashing loss to take good use of the constructed semantic similarity matrix to generate distinguished hash codes.
		\item Extensive experiments on three widely used datasets show that the proposed UHSCM outperforms state-of-the-art unsupervised baselines on image retrieval tasks.
	\end{itemize}
	
	\section{Related Work}
	\subsection{Vision-Langue Pre-training Models}
	Due to the success of BERT \cite{devlin2018bert} in NLP and ViT \cite{dosovitskiy2020image} in computer vision, more and more researchers pay attention to exploring visual-language pre-training (VLP) models \cite{li2019visualbert,su2019vl,tan2019lxmert,zhou2021learning,radford2021learning,jia2021scaling,li2021align}. With the contrastive learning, these VLP models are trained to align the semantic relationships between massive image-text pairs which are easily collected from the Internet. Benefiting from such a broader and cheaper source of data, the VLP models have shown strong power in visual representation learning. For example, the recent CLIP \cite{radford2021learning} and ALIGN \cite{jia2021scaling} employ a contrastive learning strategy on a huge amount of noisy image-text pairs, achieving surprising results on a large number of vision tasks. VATT \cite{akbari2021vatt} extends the contrastive learning from the image domain to the video domain and aligns the video frames, audios, and texts. ALBEF \cite{li2021align} proposes a new framework for vision-language representation learning which first aligns the unimodal image representation and text representation before fusing them with a multimodal encoder. 
	Moreover, motivated by CLIP, CoOp \cite{zhou2021learning} further improves the training strategy by proposing a continuous prompting optimization method to achieve better performance on visual classification tasks. 
	
	\subsection{Image Hashing}
	Depending on whether supervised information is needed in the training phase, the hashing can be roughly divided into two categories: supervised and unsupervised hashing methods. Please refer to \cite{wang2015learning,luo2020survey} for a comprehensive survey.
	
	Supervised hashing methods \cite{li2016feature,cao2017hashnet,tu2021weighted,huang2019accelerate,tu2021partial,cao2018deep,yuan2020central} learn hash functions by using not only the data representation but also the label information in the training phase. A mass of methods in this category have been proposed, such as Central Similarity Quantization (CSQ) \cite{yuan2020central} and Partial-Softmax Loss based Deep Hashing (PSLDH) \cite{tu2021partial}.  CSQ first utilizes the Hadamard matrix to generate hash centers for each label, and then pushes the hash codes of images to be close around their hash centers. PSLDH first uses category information with a bit-balance constraint to generate semantic hash codes for each category, and then it leverages the category hash codes as supervision information to guide the learning of the image hashing network by minimizing a Partial-SoftMax loss function.
	
	The unsupervised hashing methods can be divided into traditional unsupervised hashing methods and deep unsupervised hashing methods. The traditional unsupervised hashing methods \cite{weiss2009spectral, liu2014discrete,liu2012compact,gong2012iterative, liu2011hashing,yu2014circulant} use hand-crafted features and shallow hash functions to obtain binary hash codes. For example, Anchor Graph Hashing (AGH) \cite{liu2011hashing} proposes a sparse low-rank graph by introducing a set of anchors to speed up the construction of the graph to learn hash codes. Limited by the hand-crafted features and shallow hash functions, it is hard for the traditional shallow hashing methods to generate high-quality hash codes for complex and high dimensional real-world data.  Hence, plenty of  deep unsupervised hashing methods \cite{lin2016learning,yang2018semantic,tu2018object,yang2019distillhash,lin2021deep,tu2020mls3rduh,luo2020cimon} are recently proposed, which learn deep hashing networks to generate hash codes for images. For example, Semantic structure-based unsupervised deep hashing (SSDH) \cite{yang2018semantic} constructs semantic structures based on a Gaussian estimation to guide hashing network learning. Object Detection based Deep Unsupervised Hashing (ODDUH) \cite{tu2018object} first utilizes an object detection model which is pre-trained on a labeled dataset to obtain the pseudo label information of images, then based on the pseudo label information, it construct a similarity matrix for the training set to guide the learning of hashing network. MLS$^3$RDUH \cite{tu2020mls3rduh} learns the hashing network by utilizing manifold and cosine similarity between image features to reconstruct the local similarity structure.
	
	Although lots of unsupervised hashing methods have been proposed to improve the quality of similarity matrix of the training set, these constructed matrices are still not good enough because they only depend on the cosine similarity between the image features but ignore the high-level abstract semantic concepts contained in images. Hence, in this paper we propose UHSCM which leverages a VLP model with strong a generalization ability to mine the concept information for images to construct a high-quality similarity matrix.

	\begin{figure*}[tb]
		\centering
		\includegraphics[width=0.98\textwidth]{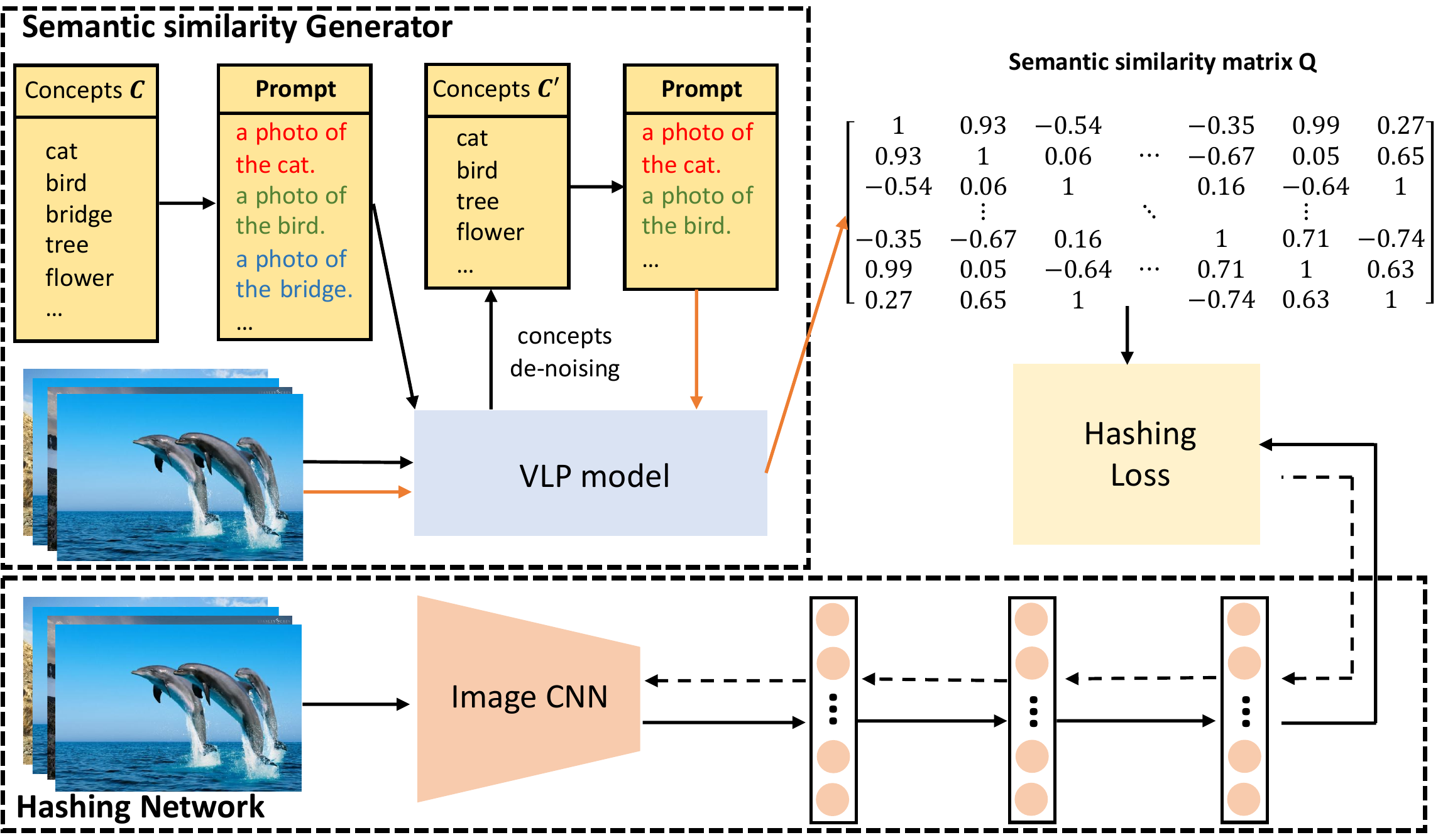}
		\caption{The architecture of UHSCM. The solid arrows indicate forward-propagation, and the dotted arrows indicate back-propagation.}
		\label{fig_architecture}
	\end{figure*}

	\section{Proposed Method}
	\subsection{Problem Definition}
	Suppose that a dataset has $n$ images $\boldsymbol{X} = \{\boldsymbol{x}_i\}_{i=1} ^n$, and $\boldsymbol{x}_i$ denotes the $i^{th}$ image. The goal of unsupervised hashing is to learn a hashing network which maps an image $\boldsymbol{x}_i$ into a similarity-preserving hash code $\boldsymbol{b}_i \in \{-1,1\}^k$ where $k$ is the length of hash codes.
	
	\subsection{Design Overview}
	As shown in Figure \ref{fig_architecture}, UHSCM mainly consists of a semantic similarity generator and a hashing network. In the semantic similarity generator, a VLP model is first used to denoise the set of randomly collected concepts according to the training images. Next, the VLP model is used again to mine the concept distributions of images, and then according to the mined concept distributions, a high-quality semantic similarity matrix of the training set is generated.  
	
	Moreover, for the hashing network,  it is a VGG19 model with the last layer replaced by a $k$-dimensional fully-connection layer and $tanh(\cdot)$ is used as the activation function for the last layer. $k$ is the length of hash codes. In the training phase, with the generated similarity matrix as guiding information, the hashing network will be optimized by a novel hashing loss to map the images $\boldsymbol{X} = \{\boldsymbol{x}_i\}_{i=1} ^n$ into hash codes $\boldsymbol{Z} = \{\mathcal{H}(\boldsymbol{x}_i; \boldsymbol{W})\}_{i=1} ^n= \{\boldsymbol{z}_i\}_{i=1} ^n\in[-1,1]^{k \times n}$, where $\mathcal{H}=(\cdot;\boldsymbol{W})$ denotes the hashing network, and $\boldsymbol{W}$ is the set of parameters of the hashing network. Then, followed by a $sgn(\cdot)$ function, the binary hash codes of images can be obtained as $\boldsymbol{B} = \{sgn(\boldsymbol{z}_i)\}_{i=1} ^n= \{\boldsymbol{b}_i\}_{i=1}^{n}\in\{-1,1\}^{k \times n}$. $sgn(\cdot)$ is an element-wise sign function which returns $1$ if the input is positive and returns $-1$ otherwise.
	% \usepackage{multirow}
	
	\subsection{Semantic Similarity Matrix Generator}	
	
	\subsubsection{Semantic Concept Mining}
	First, we randomly collect a set of concepts $\boldsymbol{C}=\{\boldsymbol{c}_i\}_i^{m}$, where $\boldsymbol{c}_i$ denotes the $i^{th}$ concept, and $m$ is the total number of concepts. For example, we can directly use the 81 classes defined in NUS-WIDE as the set of concepts. Next, for each concept, we use the following prompt template "a photo of the {$\boldsymbol{c}_i$}" to construct its corresponding text $\boldsymbol{t}_i$. Then, for an image $\boldsymbol{x}_i$, we use the VLP model to calculate an image-text similarity score vector $\boldsymbol{s}_i \in [0, 1]^{m}$, where the $j^{th}$ item of $s_{ij}$ denotes the similarity score between $\boldsymbol{x}_i$ and the $j^{th}$ concept based text  $\boldsymbol{t}_j$:
	\begin{equation}
		\begin{aligned}
		s_{ij} = \mathcal{F}_{VLP}(\boldsymbol{x}_i, \boldsymbol{t}_j; \boldsymbol{\Theta}),
		\end{aligned}
		\label{floss}
	\end{equation}
	where $\mathcal{F}_{VLP}$ denotes the pre-trained VLP model, and $\boldsymbol{\Theta}$ is the set of well-learned parameters of the VLP model. Based on the similarity score vector $\boldsymbol{s}_i$, the concept distribution of image $\boldsymbol{x}_i$ can be defined as $\boldsymbol{d}_{i}$: 
	\begin{equation}
		\begin{aligned}
			d_{ij} = \frac{e^{\tau s_{ij}}}{\sum\limits_{k = 1}^m e^{\tau s_{ik}}},
		\end{aligned}
	\end{equation}
	where $\tau$ represents the temperature parameter, and how to set its value will be described in detail in Subsection \ref{shp}; $d_{ij}$, the $j^{th}$  item of  $\boldsymbol{d}_{i}$, denotes the probability of image $\boldsymbol{x}_{i}$ containing the $j^{th}$ concept $\boldsymbol{c}_j$.  The larger value of $d_{ij}$, the higher probability of image $\boldsymbol{x}_{i}$ containing the $j^{th}$ concept $\boldsymbol{c}_j$. 
	
	Roughly, the semantic similarity $a_{ij}$ between image  $\boldsymbol{x}_{i}$ and  $\boldsymbol{x}_{j}$ can be directly calculated by their corresponding concept distributions $\boldsymbol{d}_{i}$ and $\boldsymbol{d}_{j}$:
	\begin{equation}
		\begin{aligned}
			a_{ij} = \frac{\boldsymbol{d}_{i}^{T}\boldsymbol{d}_{j}}{\left\|\boldsymbol{d}_{i}\right\|_2\left\|\boldsymbol{d}_{j}\right\|_2},
		\end{aligned}
		\label{acal}
	\end{equation}
	where $\left\|\cdot\right\|_2$ denotes $l_2$ norm.
	
	However, such a way of constructing semantic similarity is not a good choice for the following reason. Since the set of concepts is randomly collected, some of them may not be appropriate for the image dataset, making these concepts useless for distinguishing the images.	If we do not eliminate these concepts,  they will become noisy and harm the quality of the similarities constructed based on the concept distributions. For example, suppose that a concept $\boldsymbol{c}_i$ is such a noise that all the images in the dataset do not contain it. But the VLP model may misjudge two dissimilar images both containing the concept $\boldsymbol{c}_i$ on a high probability, so the two dissimilar images will be mistaken for similar which will misguide the training of the hashing model and harm the retrieval performance. Hence, we propose a way to denoise the set of concepts to make the constructed semantic similarities more high-quality.
	
	\subsubsection{Semantic Concept  Denoising}
	\label{scd}
	With the above calculated concept distributions $\boldsymbol{D} = \{\boldsymbol{d}_i\}_{i=1}^n$ of all the images $\boldsymbol{X}$,  then for a concept $\boldsymbol{c}_i$, we can calculate its frequency $f(\boldsymbol{c}_i)$ where it denotes the number of images containing the concept $\boldsymbol{c}_i$ in the highest probability:
	\begin{equation}
		\begin{aligned}
			f(\boldsymbol{c}_i) = \sum\limits_{k = 1}^n I(j=i | \mathop{\arg\max}_{j} \boldsymbol{d}_{kj}),
		\end{aligned}
	\label{f}
	\end{equation}
	where $I(\cdot)$ is a conditional function which returns 1 when the condition is true otherwise returns 0.
	
	Intuitively, for the concept $\boldsymbol{c}_i$, if its frequency $f_i$ is very large or small, i.e., most or only a little bit of images contain it in the highest probability, the concept is a noise, which is useless for distinguishing these images and even will harm the quality of the constructed similarities. Hence, it will be better to discard this concept. Based on that,  we  define a conditional function to determine whether the concept $\boldsymbol{c}_i$ needs to be discarded:
	\begin{equation}
		Discard(\boldsymbol{c}_i) = \left\{
		\begin{aligned}
			0,\ \ \ & if\  0.5\frac{n}{m} \le f(\boldsymbol{c}_i) \le 0.5n;\\
			1,\ \ \  &otherwise,
		\end{aligned}
	\label{d}
	\right .
	\end{equation}
	where $n$ is the number of training images, and $m$ is the total number of concepts. When the frequency $f(\boldsymbol{c}_i)$ is larger than $0.5n$, more than half of  images contain the concept $\boldsymbol{c}_i$ with the highest probability.  It means that more than half of images are defined as similar to each other which will compromise the distinguishability of the generated hash codes; when $f(\boldsymbol{c}_i)$ is smaller than $0.5\frac{n}{m}$, the number of images containing the concept $\boldsymbol{c}_i$ is even smaller than half of the average number of images each concept contained, which is too small then such a concept may not be suitable for this dataset. Hence, in these two cases, the concept $\boldsymbol{c}_i$ should be discarded, i.e., $Discard(\boldsymbol{c}_i)=1$; in the other cases, i.e., $0.5\frac{n}{m} \le f(\boldsymbol{c}_i) \le 0.5n$, it means that the concept may be useful for distinguishing the images, then $Discard(\boldsymbol{c}_i)=0$.
	
	By keeping the concept $\boldsymbol{c}_i$ whose $Discard(\boldsymbol{c}_i)=0$ and discarding the one whose $Discard(\boldsymbol{c}_i)=1$, we can denoise the original set of concepts $\boldsymbol{C}$ to obtain a cleaner concept set $\boldsymbol{C}'={\{\boldsymbol{c}'}_i\}_i^{m'}$, where $m'$ denotes the number of retained useful concepts. 
	
	Similar to the calculation of $\boldsymbol{D}=\{\boldsymbol{d}\}_{j=1}^n$,  with the set of concepts $\boldsymbol{C}'={\{\boldsymbol{c}'}_i\}_i^{m'}$, we use the VLP model through prompt engineering again to calculate the denoised concept distributions $\boldsymbol{D}'=\{\boldsymbol{d}'\}_{j=1}^n$ for the images $\boldsymbol{X}$. Finally, we can calculate the high-quality similarity matrix $\boldsymbol{Q}$:
	\begin{equation}
		\begin{aligned}
		q_{ij} = \frac{{\boldsymbol{d}'}_{i}^{T}{\boldsymbol{d}_{j}'}}{\left\|{\boldsymbol{d}_{i}'}\right\|_2\left\|{\boldsymbol{d}_{j}'}\right\|_2},
		\end{aligned}
	\end{equation}
	where $q_{ij}$ is the $i^{th}$ row $j^{th}$ column of $\boldsymbol{Q}$ and denotes the semantic similarity between image $\boldsymbol{x}_i$ and image $\boldsymbol{x}_j$.
	
	\subsection{Learning to Hash}
	Now, by taking the semantic similarity matrix $\boldsymbol{Q}$ as guiding information, we will train the hashing model to map the images into hash codes with the constructed semantic similarity preserved, i.e., when $q_{ij}$ is so large that image $\boldsymbol{x}_i$ and image $\boldsymbol{x}_j$ are similar, the Hamming distance between their corresponding hashing codes $\boldsymbol{b}_i$ and $ \boldsymbol{b}_j$, which is defined as $H_d(\boldsymbol{b}_i, \boldsymbol{b}_j)=\frac{1}{2}(k - \boldsymbol{b}_{i}^T\boldsymbol{b}_j)$,  should be small, otherwise $H_d(\boldsymbol{b}_i, \boldsymbol{b}_j)$ should be large.  To achieve this goal, the objective function $\mathcal{L}$ of our hash model is defined as follows: 
	\begin{equation}
		\begin{aligned}
			\mathcal{L}_s = \frac{1}{n^2}\sum\limits_{i = 1}^n\sum\limits_{j = 1}^n(h_{ij} - q_{ij})^2,
		\end{aligned}
	\end{equation}
	\begin{equation}
		\begin{aligned}
			\mathcal{L}_c = \frac{1}{n}\sum\limits_{i = 1}^n\sum\limits_{j \in \Psi_i}  \frac{1}{|\Psi_i|} \frac{e^{h_{ij}/\gamma}}{e^{h_{ij}/\gamma} + \sum\limits_{l\in \Phi_i}e^{h_{il}/\gamma}} ,
		\end{aligned}
	\label{cl}
	\end{equation}
	\begin{equation}
		\begin{aligned}
			\mathcal{L} =  \mathcal{L}_s + & \alpha \mathcal{L}_c,
		\end{aligned}
	\end{equation}
	where $\alpha$ and $\gamma$ are the hyper-parameters; $h_{ij}$ denotes the Hamming similarity between the hash codes $\boldsymbol{b}_i$ and $ \boldsymbol{b}_j$ which is defined as $h_{ij} = \frac{\boldsymbol{b}_{i}^{T}\boldsymbol{b}_{j}}{\left\|\boldsymbol{b}_{i}\right\|_2\left\| \boldsymbol{b}_{j} \right\|_2} = \frac{1}{k}\boldsymbol{b}_i^T\boldsymbol{b}_j$. Moreover, $\Psi_i = \{j|q_{ij}\geq\lambda\}$ is the set of indexes of images to which the image $\boldsymbol{x}_i$ should be similar, and $\lambda$ is a hyper-parameter. $\Phi_i = \{j|q_{ij}\textless\lambda\}$ is the set of indexes of images which are not in  $\Psi_i$
	
	Specifically, the first item $\mathcal{L}_s$ is a widely used $l_2$ loss, which is used to make the learned hash codes preserve the constructed semantic similarity matrix $\boldsymbol{Q}$ well. It can be found that by minimizing the $\mathcal{L}_s$, when $q_{ij}$ is large, the value of $h_{ij}$, i.e., $\frac{1}{k}\boldsymbol{b}_{i}^T\boldsymbol{b}_j$ will be large, then the Hamming distance $H_d(\boldsymbol{b}_i, \boldsymbol{b}_j)$ will be small; otherwise, $H_d(\boldsymbol{b}_i, \boldsymbol{b}_j)$ will be large.
	
	Moreover, the second item $\mathcal{L}_c$ is a modified contrastive loss proposed by us. It can help our hash model further exploit the semantic similarity matrix $\boldsymbol{Q}$ to generate distinguished hash codes. The proposition of $\mathcal{L}_c$ is inspired by the tremendous success of contrastive learning. Although the contrastive loss has been introduced to optimize the hashing model by the CIB \cite{qiu2021unsupervised}, it only treats the different views of the same image as similar pairs but does not leverage the constructed semantic similarities to find useful similar data pairs to further improve the retrieval performance. The contrastive loss $\mathcal{J}_c$ defined in the CIB \cite{qiu2021unsupervised} is as follows:
	 \begin{equation}
		\begin{aligned}
			\mathcal{J}_c = \frac{1}{n} \sum\limits_{i = 1}^n\frac{e^{sim(\boldsymbol{b}_{i}^{(1)}, \boldsymbol{b}_{i}^{(2)})/\gamma}}{e^{sim(\boldsymbol{b}_{i}^{(1)}, \boldsymbol{b}_{i}^{(2)})/\gamma} + \sum\limits_{l\in {1,2}}\sum\limits_{k \not= i}e^{e^{sim(\boldsymbol{b}_{i}^{(1)}, \boldsymbol{b}_{k}^{(l)})/\gamma}}} ,
			\end{aligned}
		\end{equation}
	where $\boldsymbol{b}_{i}^{(1)}$ and $\boldsymbol{b}_{i}^{(2)}$ denote the hash codes of different views of $\boldsymbol{x}_{i}$; $sim(\boldsymbol{b}_{i}^{(1)}, \boldsymbol{b}_{i}^{(2)})$ denotes the Hamming similarity between hash codes.  By minimizing the loss $\mathcal{J}_c$, it just only makes the Hamming similarity between hash codes of different views of the same images higher than that between hash codes of any two different images, no matter whether the two images are similar. It means that the contrastive loss $\mathcal{J}_c$ ignores an amount of useful similarity information between different images. Hence, in our paper, we propose a novel modified contrastive loss $\mathcal{L}_c$ defined as Formula (\ref{cl}). Different from $\mathcal{J}_c$, without generating different views of images through data augmentation, our modified contrastive loss $\mathcal{L}_c$ directly leverages the semantic similarity matrix $\boldsymbol{Q}$ to construct the similar data pairs. Specifically, in $\mathcal{L}_c$, when $\boldsymbol{q}_{ij} \geq \lambda$, the image $\boldsymbol{x}_j$ is treated similar to image $\boldsymbol{x}_i$, i.e., the index $j$ belongs to the set $\Psi_i$ of image $\boldsymbol{x}_i$. Then, after minimizing $\boldsymbol{L}_c$, for each $j$ in $\Psi_i$, the Hamming similarity between $\boldsymbol{b}_i$ and $\boldsymbol{b}_j$ will be larger than that between $\boldsymbol{b}_i$ and $\boldsymbol{b}_k$ where $k \not \in \Psi_i$, Hence, it makes the generated hash codes more distinguished.
	
	Furthermore, as the each hash code $\boldsymbol{b}_i=sign(\mathcal{H}(\boldsymbol{x}_i; \Theta))$ and the $sign(\cdot)$ function is in-differentiable at zero and the derivation of it will be zeros for a non-zero input. It means that the parameters of hashing model will not be updated with the back-propagation algorithm when minimizing the loss function $\mathcal{L}$. Thus, to ensure the parameters of our hashing model are updated, we directly utilize $\tanh(\cdot)$ to approximate the $sign(\cdot)$ function,	Then, similar to previous works \cite{song2018binary, tu2021partial}, we  further add a quantization loss to make each element of outputs of the hashing network close to ``+1" or ``-1". The final objective function can be formulated as follows:
	\begin{equation}
		\begin{aligned}
			\mathcal{L} = &\frac{1}{n^2}\sum\limits_{i = 1}^n\sum\limits_{j = 1}^n(\hat{h}_{ij} - q_{ij})^2 + \beta \frac{1}{n}\sum\limits_{i = 1}^n\left \|\boldsymbol{z}_{i} - \boldsymbol{b}_i\right \|_F^2\\
			&+ \frac{\alpha}{n}\sum\limits_{i = 1}^n\sum\limits_{j \in \Psi_i}  \frac{1}{|\Psi|} \frac{e^{\hat{h}_{ij}/\gamma}}{e^{\hat{h}_{ij}/\gamma} + \sum\limits_{l\in \Phi_i}e^{\hat{h}_{il}/\gamma}} ,
		\end{aligned}
	\label{of}
	\end{equation}
	where $\hat{h}_{ij} = \frac{\boldsymbol{z}_{i}^{T}\boldsymbol{z}_{j}}{\left\|\boldsymbol{z}_{i}\right\|_2\left\| \boldsymbol{z}_{j} \right\|_2}$; $\boldsymbol{z}_i$ denotes the hash codes of image $\boldsymbol{x}_i$, i.e., the output of hashing network with the input $\boldsymbol{x}_i$. The details of the learning procedure are shown in Algorithm \ref{alg}.
	
	\begin{algorithm}[t]
		\caption{Learning algorithm for UHSCM}
		\label{alg}
		%\textbf{Input}: Images $\boldsymbol{X}$,  the length of  hash codes $k$. A random selected set of concepts $\boldsymbol{C}$.
		%\textbf{Output}: Parameters of hashing network $\boldsymbol{W}$, hash codes $\boldsymbol{B}$.
		\begin{algorithmic}[1]
			\REQUIRE Images $\boldsymbol{X}$,  the length of  hash codes $k$, a set of randomly selected concepts $\boldsymbol{C}$.
			\ENSURE Parameters of hashing network $\boldsymbol{W}$, hash codes $\boldsymbol{B}$.
			\STATE Initialize parameters: $\boldsymbol{W}$, $\alpha$, $\beta$\, $\gamma$, $\lambda$, $k$, $o$. learning rate: $lr$, iteration number: $T$, mini-batch size  $t$ (see Section \ref{setting}).
			\STATE Calculate the concept distributions $\boldsymbol{D}$ of images $\boldsymbol{X}$ over the set of concepts $\boldsymbol{C}$ by the VLP model through prompting.
			\STATE Obtain the clean concept set $\boldsymbol{C}'$ by denoising $\boldsymbol{C}$ with the Formula (\ref{f}) and Formula (\ref{d}).
			\STATE Generate the concept distributions $\boldsymbol{D}'$ through the prompt engineering.
			\STATE Calculate the semantic similarity matrix $Q$.
			\REPEAT
			\FOR{$j=1:\frac{n}{t}$}
			\STATE Randomly sample $t$ images from database as a mini-batch.
			\STATE Generate hash code $\boldsymbol{z}_i$ with image $\boldsymbol{x}_i$ as input by the hash network.
			\STATE Update parameters of the hash network $\boldsymbol{W}$ by minimizing Formula (\ref{of}).
			\ENDFOR
			\UNTIL{Convergence}
			\STATE Generate binary image hash codes $\boldsymbol{B}$.
		\end{algorithmic}
	\end{algorithm}

	\begin{table*}[]
		\centering
		\begin{tabular}{c|cccc|cccc|cccc}
			\toprule[1.2pt]
			\multirow{2}{*}{Method} & \multicolumn{4}{c|}{CIFAR10}      & \multicolumn{4}{c|}{NUS-WIDE}  & \multicolumn{4}{c}{MIRFlickr-25K}        \\ \cline{2-13} 
			&32 bits & 64 bits&96 bits & 128 bits &32 bits & 64 bits&96 bits & 128 bits &32 bits & 64 bits&96 bits & 128 bits  \\ \hline
			LSH&0.257	&0.286	&0.346	&0.375	&0.538	&0.579	&0.636	&0.666	&0.642	&0.685	&0.701	&0.702 \\
			SH &0.327	&0.339	&0.341	&0.353	&0.612	&0.623	&0.623	&0.626	&0.660	&0.659	&0.654	&0.654 \\
			ITQ &0.442	&0.474	&0.479	&0.492	&0.719	&0.743	&0.751	&0.753	&0.763	&0.769	&0.776	&0.776  \\
			AGH &0.495	&0.491	&0.485	&0.481	&0.727	&0.733	&0.734	&0.732	&0.798	&0.786	&0.777	&0.771           \\ \hline \hline
			SSDH             &0.314	&0.331	&0.352	&0.372	&0.552	&0.596	&0.637	&0.673	&0.749	&0.752	&0.761	&0.762        \\ 
			GH             &0.456	&0.469	&0.500	&0.504	&0.684	&0.720	&0.737	&0.743	&0.744	&0.766	&0.782	&0.791          \\
			BGAN                   &0.583	&0.607	&0.604	&0.612	&0.777	&0.785	&0.790	&0.793	&0.783	&0.793	&0.803	&0.806           \\ 
			MLS$^3$RDUH           &0.540	&0.550	&0.559	&0.569	&0.776	&0.788	&0.793	&0.796	&0.814	&0.818	&0.817	&0.816            \\ 
			CIB                     &0.580	&0.599	&0.606	&0.611	&0.774	&0.782	&0.782	&0.783	&0.796	&0.808	&0.813	&0.812       \\ \hline\hline
			UHSCM                &\textbf{0.831}	&\textbf{0.850}	&\textbf{0.857}	&\textbf{0.853}	&\textbf{0.796}	&\textbf{0.810}	&\textbf{0.813}	&\textbf{0.815}	&\textbf{0.827}	&\textbf{0.834}	&\textbf{0.835}	&\textbf{0.834}  \\  \toprule[1.2pt]
		\end{tabular}
		\caption{MAPs of Hamming ranking for different numbers of hash bits on the three image datasets.}
		\label{map}
	\end{table*}
	\section{Experiments}
	In this section, to evaluate the proposed method UHSCM, extensive experiments are conducted on three commonly used datasets.
	\subsection{Datasets and Settings}
	\label{setting}
	The three datasets used for evaluation are \textit{\textbf{CIFAR10}} \cite{krizhevsky2009learning}, \textit{\textbf{MIRFlickr-25K}} \cite{huiskes2008mir} and \textit{\textbf{NUS-WIDE}} \cite{chua2009nus} which are described below.
	
	\textit{\textbf{CIFAR10}} is a popular image dataset containing 60,000 images in 10 classes, where each class contains 6,000 images 
	with size 32 × 32. We randomly sample 100 images for each class as the test set and use the remaining images as the database, 1,000 images per class from the database as the training set. 
	
	\textit{\textbf{NUS-WIDE}} dataset contains 269,648 images crawled from Flickr. Each image is annotated with one or multiple labels from 81 concept labels. To ensure sufficient samples in each class, only 195,834 images that belong to the 21 most frequent classes are selected for our experiments. Then, we randomly sample 5,000 images as the test set and use the remaining images as the database, 10,500 images from the database as the training set.
	
	\textit{\textbf{MIRFlickr-25K}}  contains 25,000 images and each image is labeled with at least one of 24 classes labels. We randomly selected 1,000 images as the test set and the remaining images as the database. In the database, we randomly sample 10,000 images as the training set.

	\begin{figure*}[t]
		\centering
		\subfigure[64 bits]{
			\begin{minipage}[t]{0.28\linewidth}
				\centering
				\includegraphics[width=\linewidth]{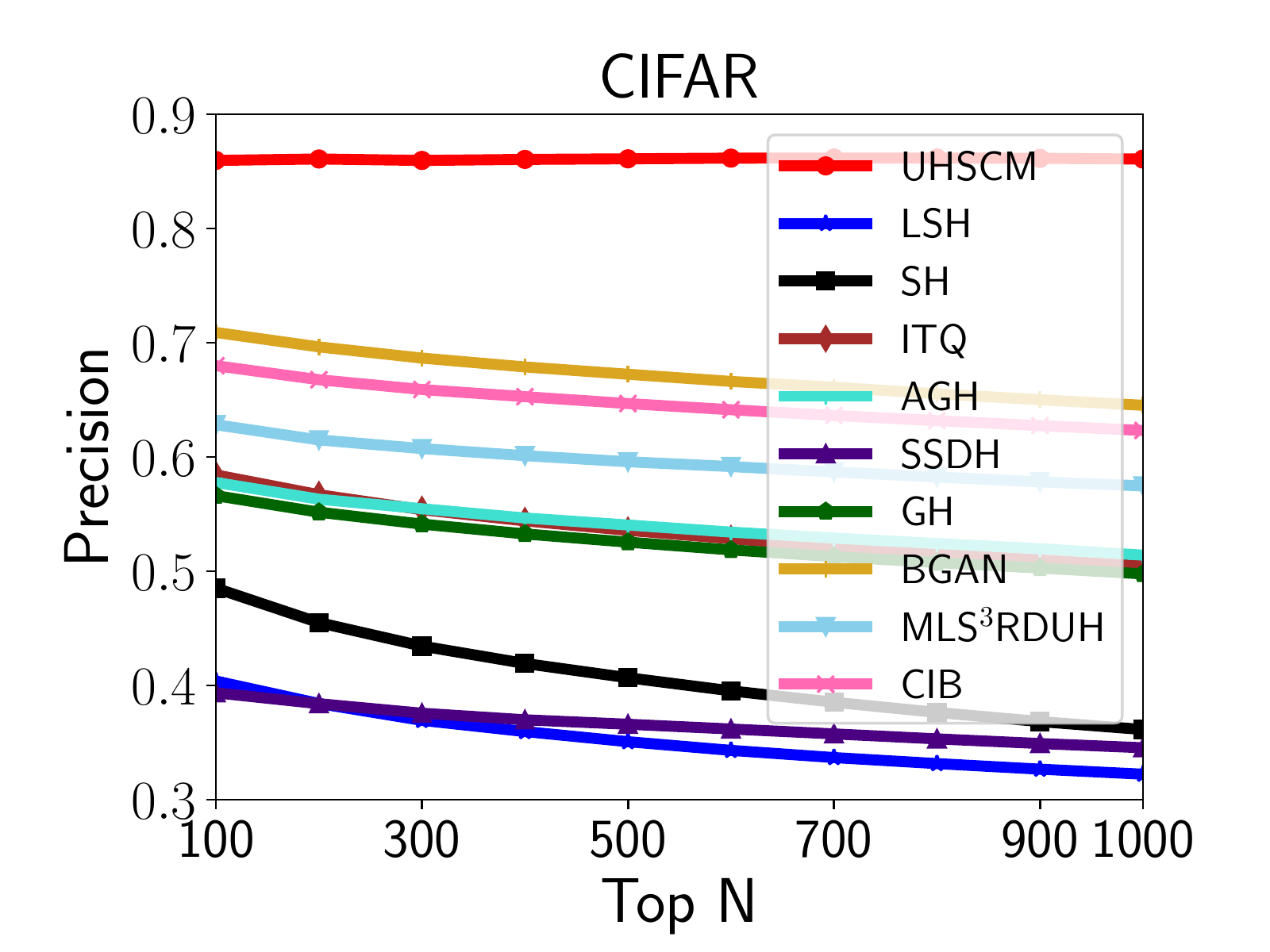}
				%\caption{fig1}
			\end{minipage}%
		}%
		\subfigure[128 bits]{
			\begin{minipage}[t]{0.28\linewidth}
				\centering
				\includegraphics[width=\linewidth]{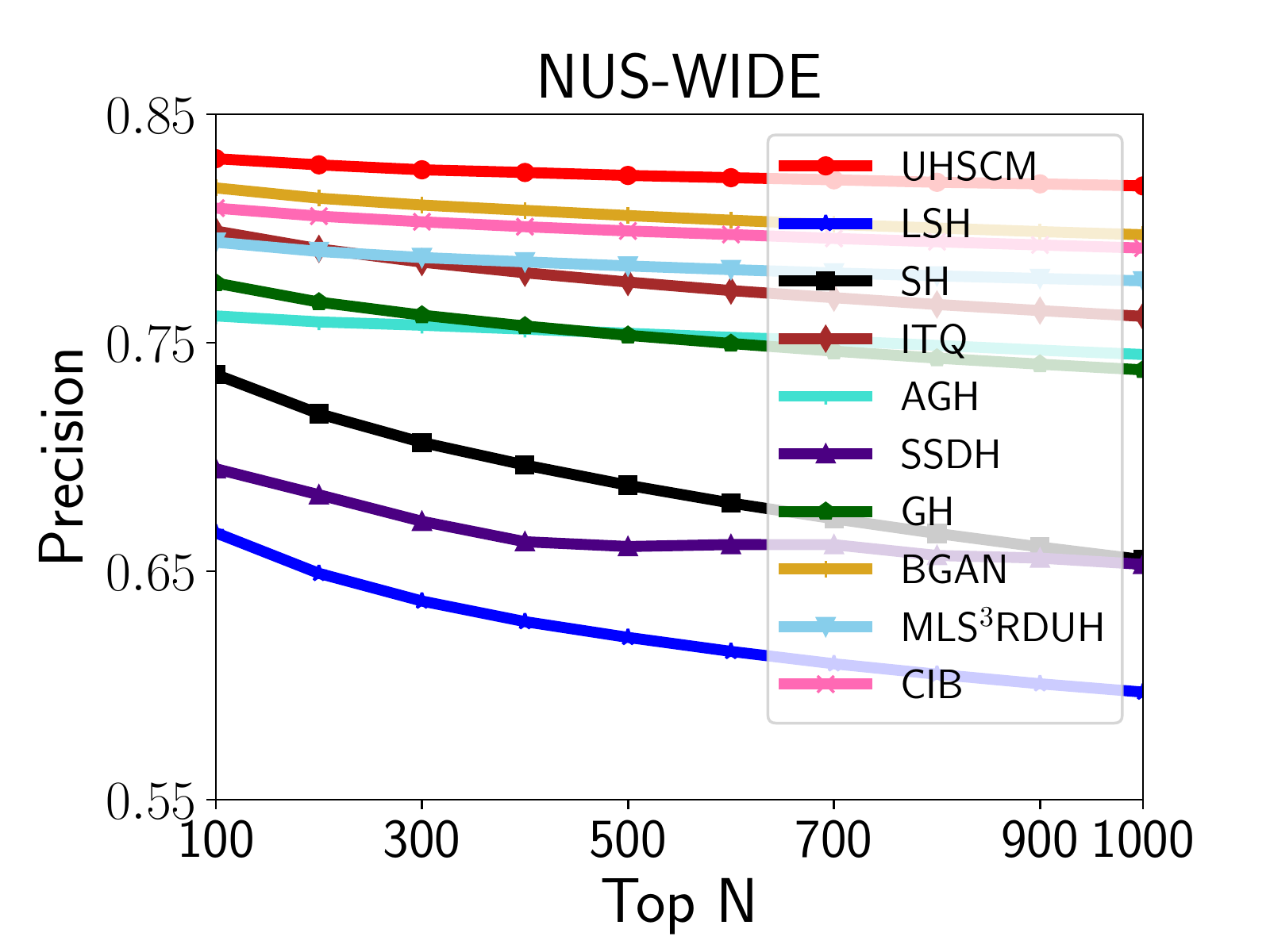}
				%\caption{fig1}
			\end{minipage}%
		}%
		\subfigure[64 bits]{
			\begin{minipage}[t]{0.28\linewidth}
				\centering
				\includegraphics[width=\linewidth]{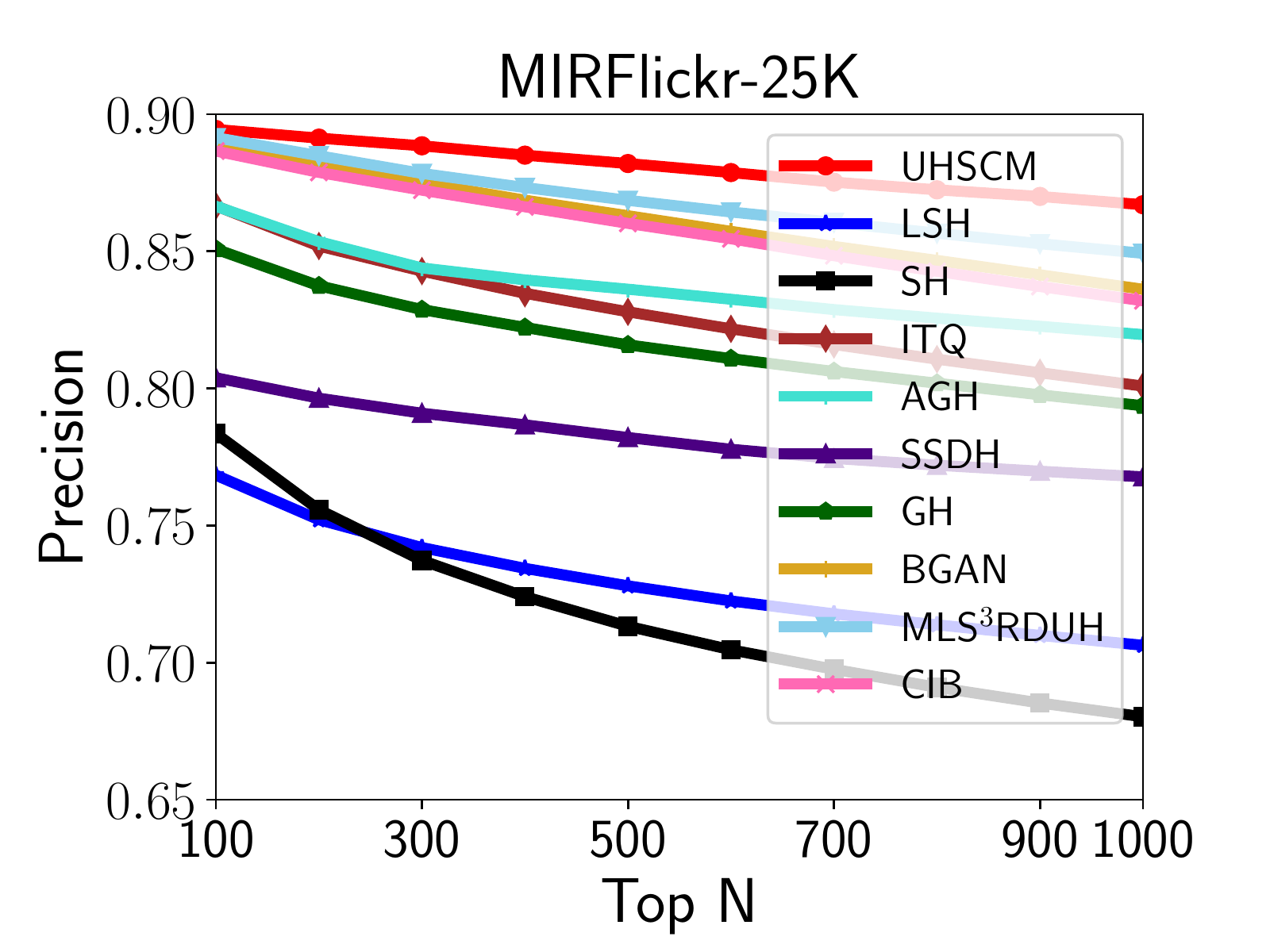}
				%\caption{fig1}
			\end{minipage}%
		}%
		\quad
		\subfigure[128 bits]{
			\begin{minipage}[t]{0.28\linewidth}
				\centering
				\includegraphics[width=\linewidth]{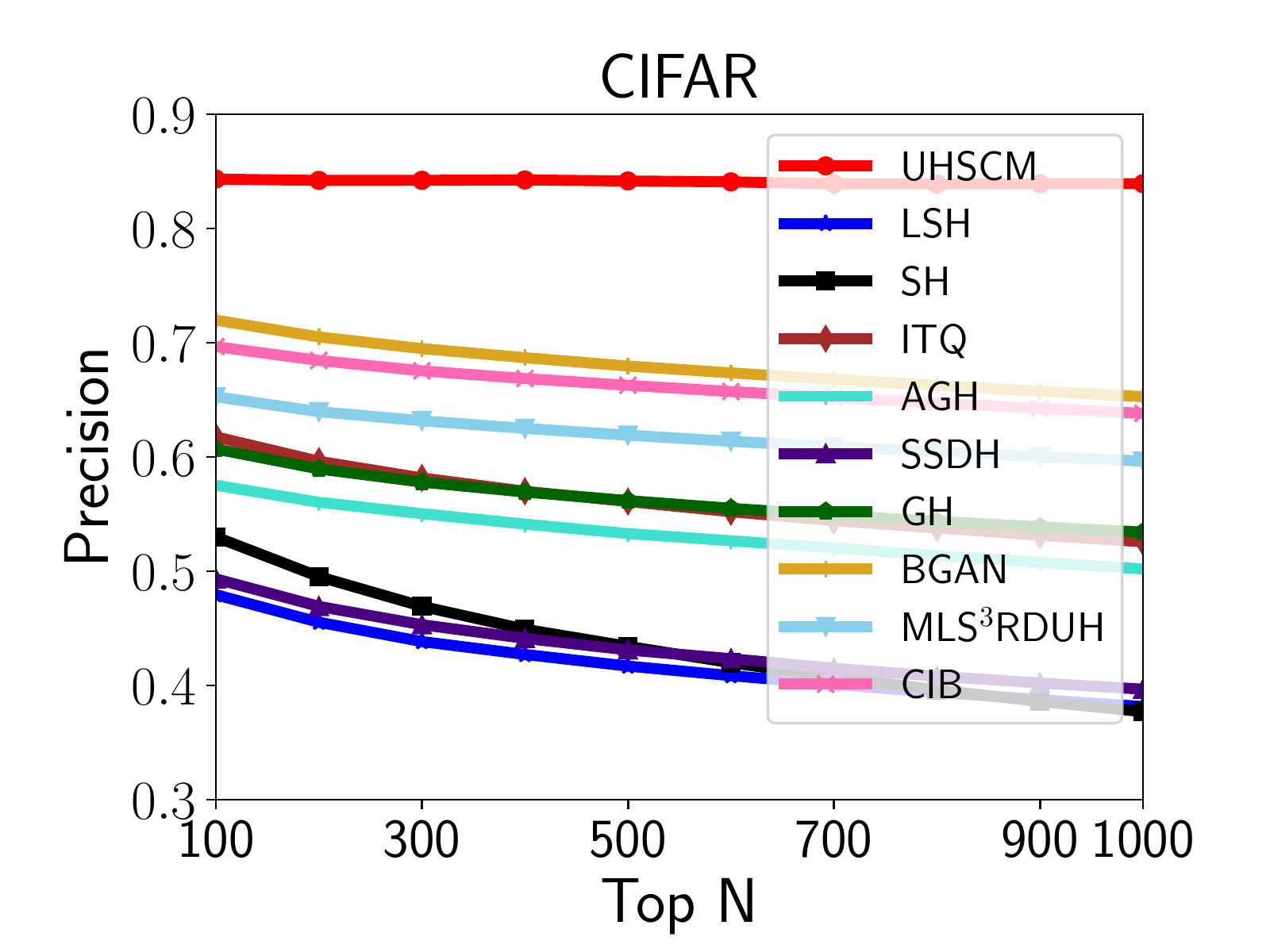}
				%\caption{fig1}
			\end{minipage}%
		}%
		\subfigure[64 bits]{
			\begin{minipage}[t]{0.28\linewidth}
				\centering
				\includegraphics[width=\linewidth]{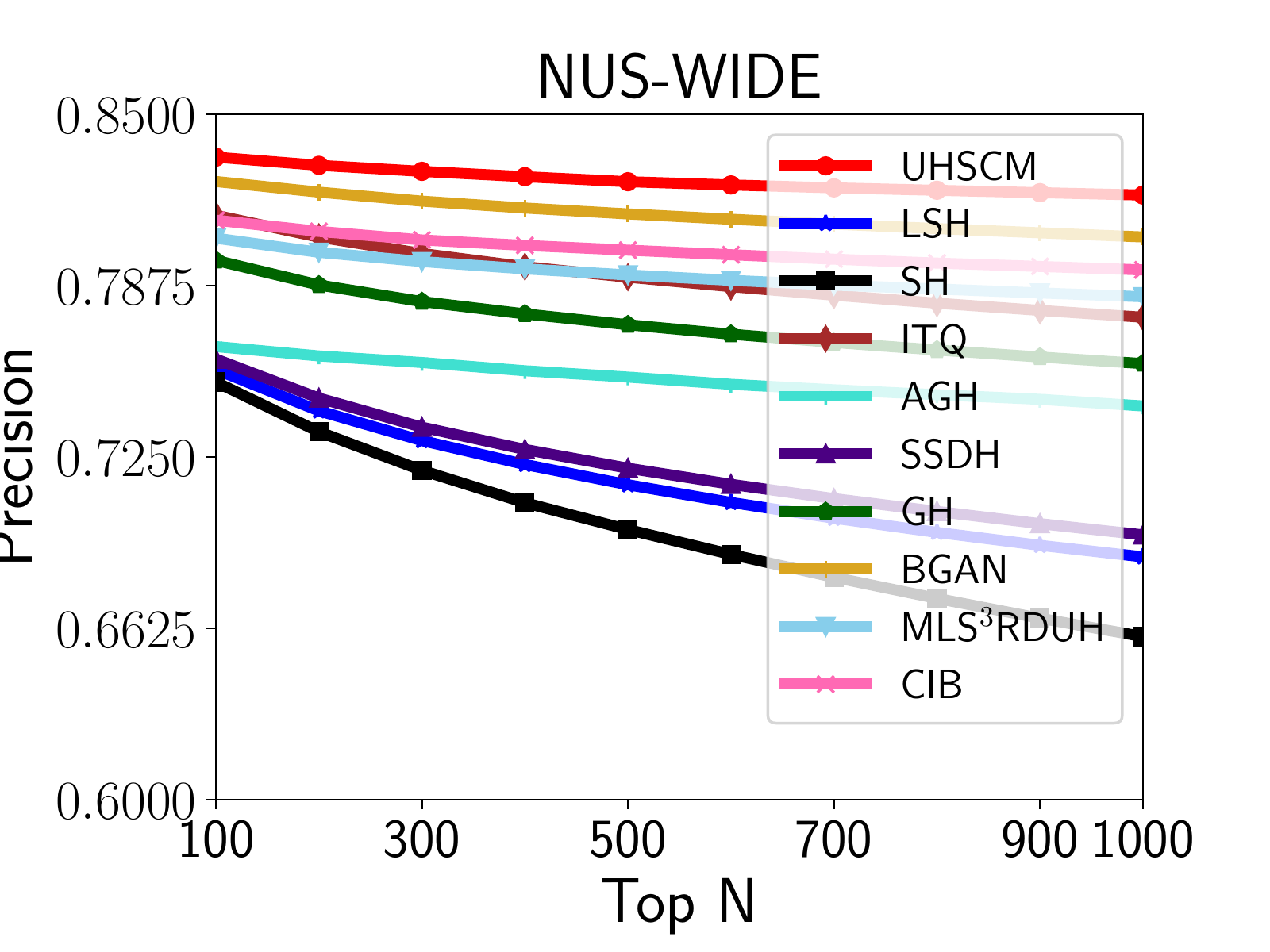}
				%\caption{fig1}
			\end{minipage}%
		}%
		\subfigure[128 bits]{
			\begin{minipage}[t]{0.28\linewidth}
				\centering
				\includegraphics[width=\linewidth]{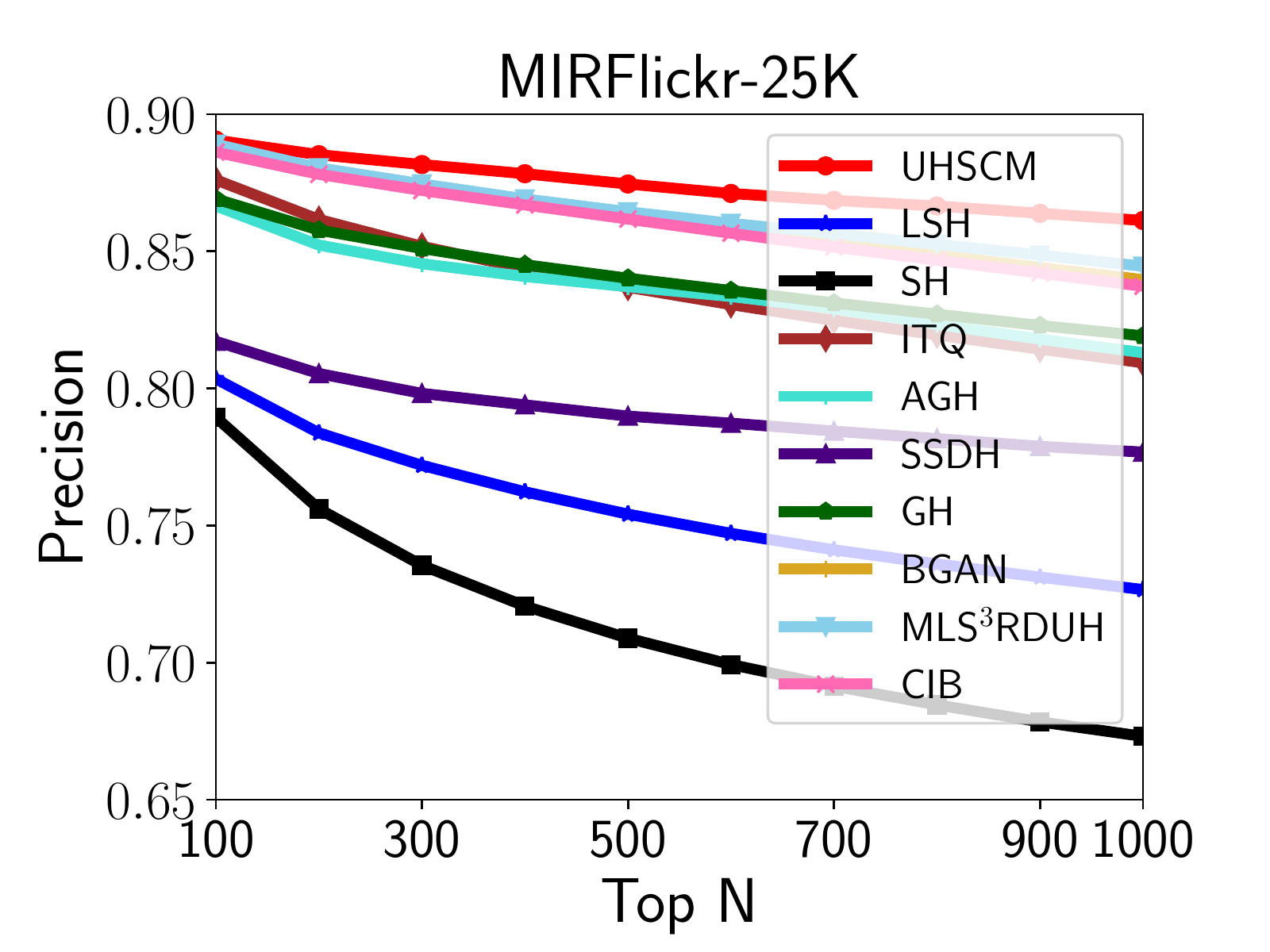}
				%\caption{fig1}
			\end{minipage}%
		}%
	\vspace{-0.3cm}
		\caption{Precision@N curves on the three datasets}
		\vspace{-0.3cm}
		\label{fig_p}
	\end{figure*}

	As our proposed method UHSCM is an unsupervised method, we compare it with nine classical and state-of-the-art unsupervised hashing methods: four traditional shallow unsupervised methods LSH \cite{gionis1999similarity}, SH \cite{weiss2009spectral}, ITQ \cite{gong2012iterative} and AGH \cite{liu2011hashing}; five deep unsupervised hashing methods: UTH \cite{huang2017unsupervised}, SSDH \cite{yang2018semantic}, GH \cite{su2018greedy}, BGAN \cite{song2018binary}, MLS$^3$RDUH \cite{tu2020mls3rduh}, CIB \cite{qiu2021unsupervised}. For fair comparison, we adopt the VGG19 architecture \cite{simonyan2014very} for all the deep hashing methods, and the inputs of the deep hashing methods are the raw images.  Moreover, we extract 4,096-dimensional deep features by VGG19 model which is pre-trained on ImageNet \cite{russakovsky2015imagenet} dataset as the inputs of the four shallow hashing methods.
	
	In our implementation of UHSCM, we utilize the VGG19 architecture \cite{simonyan2014very} and implement it based on Pytorch framework. We use all the 81 categories of NUS-WIDE directly as the original concepts for all three experimental datasets CIFAR10, NUS-WIDE and MIRFlickr-25k. Note that when conducting the hashing retrieval experiments on the NUS-WIDE dataset, only the images belonging to the 21 most frequent categories are selected. This means that the collected original concepts for the hashing retrieval experiments on the NUS-WIDE dataset still contain many noise concepts.
	Moreover, the VLP model used in our method is the pre-trained CLIP\footnote{https://github.com/openai/CLIP} model \cite{radford2021learning}. The parameters in the first eighteen layers of hashing model are initialized with the parameters of the first eighteen layers of VGG19  model which is pre-trained on ImageNet, and the parameters in the nineteen layer of hashing model are initialized by Xavier initialization \cite{glorot2010understanding}. We use mini-batch stochastic gradient descent (SGD) with 0.9 momentum and the learning rate is fixed to $0.006$. We fix the mini-batch size of images as 128 and the weight decay parameter as $10^{-5}$.

	\subsection{Evaluation Protocol} 
	For hashing based retrieval tasks, Hamming ranking and hash lookup are two widely used retrieval protocols to evaluate the performance of hashing methods \cite{tu2022deep,tu2020deep,qiu2021unsupervised}. In our experiments, similar to \cite{tu2021partial}, we evaluate the retrieval quality based on three evaluation metrics: Mean Average Precision (\textbf{MAP}), Precision curves with respect to the number of top N returned results (\textbf{P@n}), Precision-Recall curves (\textbf{PR}). MAP, P@n are used to measure the accuracy of the Hamming ranking protocol. PR curve is used to evaluate the accuracy of the hash lookup protocol. Moreover, for the MAP, P@n, and PR curve, the image $\boldsymbol{x}_i$ and image $\boldsymbol{x}_j$ will be defined as a similar pair if $\boldsymbol{x}_i$ and $\boldsymbol{x}_j$ share at least one common label. Otherwise, they will be defined as a dissimilar pair.  
	
	Specifically, given a query datapoint, the Average Precision (AP) score of top n retrieved datapoints is defined as:
	\begin{equation}
		AP = \sum_{i=1}^n \frac{I(i)}{N}\sum_{j=1}^i \frac{I(j)}{i},
	\end{equation}
	where $I(i)$ is an indicator function, if the $i^{th}$ retrieved image is relevant to the query, $I(i)=1$ ; otherwise $I(i) = 0$. $N$ represents the number of relevant images in the returned top $n$ datapoints. Then, the Mean Average Precision (MAP)  is defined as the average of APs for all queries. Moreover, for all the three datasets, we set n as 5000.

	\subsection{Experimental Results}
	
	\subsubsection{Hamming Ranking Protocol}
	Table \ref{map} shows the MAP  results of all baselines and UHSCM on CIFAR10, NUS-WIDE, and MIRFlickr-25K datasets, respectively. The P@N curves on 64 and 128bits over the three datasets are shown in Figure \ref{fig_p} . From the table and figures, it can be observed that our method outperforms all state-of-the-art baselines on both the two evaluation metrics.  For instance, compared with the latest baseline CIB, the MAP results of our proposed UHSCM have an average increase of 24.8\%, 2.8\% and 2.5\% on datasets CIFAR10, NUS-WIDE and MIRFlickr-25K, respectively. Compared with the best competitor MLS$^3$RDUH on the NUS-WIDE dataset, the MAP results of UHSCM have an average increase of 2.0\%.  Moreover, as shown in Figure \ref{fig_p}, the P@N curves of our method are the highest, and especially the results on CIFAR10 dataset, the Precision@N curves of our proposed UHSCM are greatly larger than all the baselines.
	These results indicate that the hash codes generated by our proposed method UHSCM can preserve  more semantic similarity information than state-of-the-art baselines. 
	
	In addition, our proposed UHSCM achieves a large performance improvement on the CIFAR10 dataset compared to the baselines, but the performance improvements on the other two datasets are not as large as on the CIFAR10 dataset. There are two reasons for this: 1) The CIFAR10 is a single-label dataset where it is easy to mine the concept to define high-quality semantic similarity matrix, while the NUS-WIDE and MIRFlickr-25K datasets are multi-label datasets that contain many objects in each image, thus it is difficult to mine the comprehensive and accurate concepts. Therefore, our method can achieve better performance improvement on the CIFAR10 dataset. 2) The performances of existing methods on CIFAR10 are very poor, so it is easy for our method to improve the performance.

	\begin{figure*}[t]
		\centering
		\subfigure[64 bits]{
			\begin{minipage}[t]{0.28\linewidth}
				\centering
				\includegraphics[width=\linewidth]{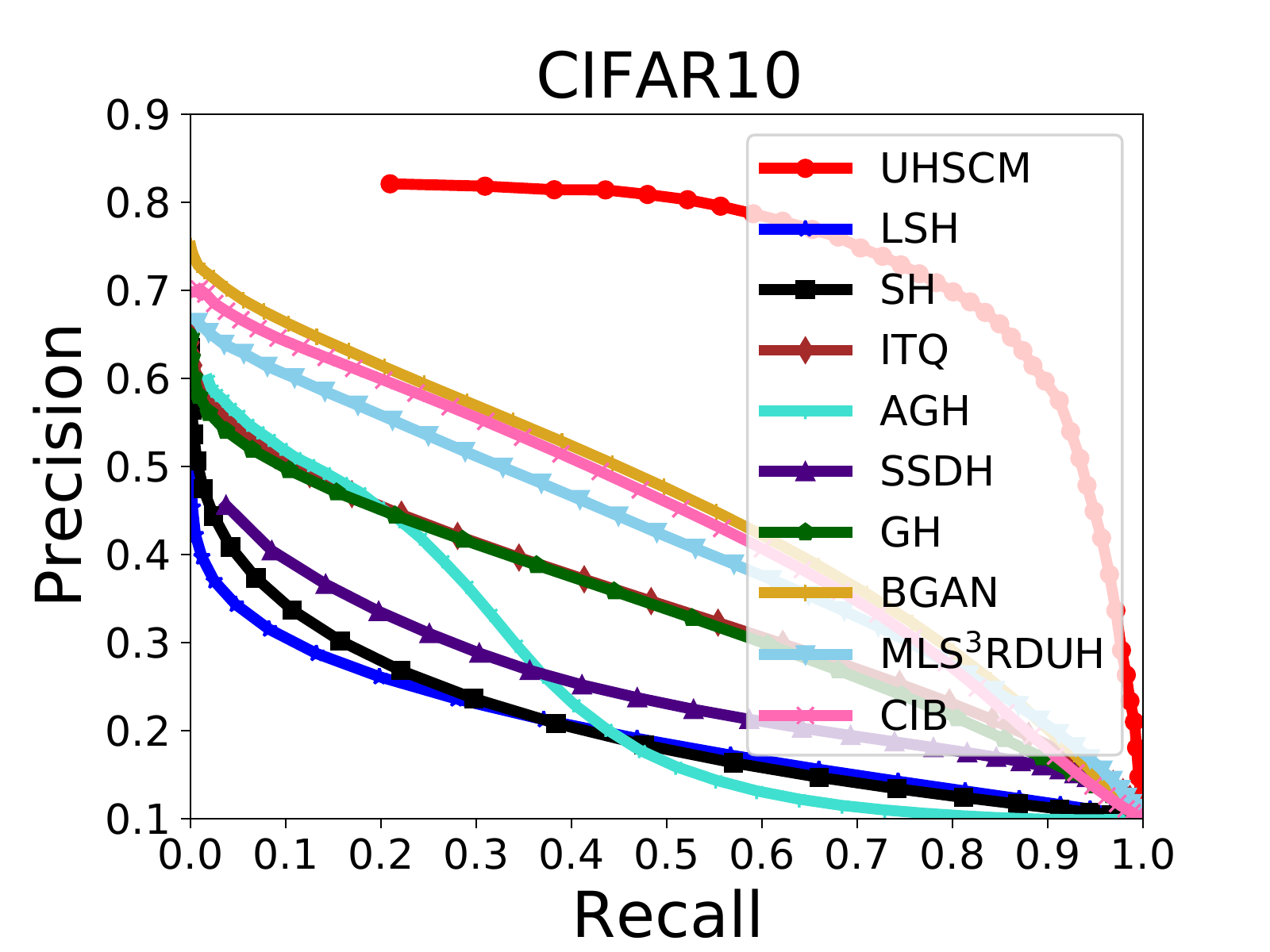}
				%\caption{fig1}
			\end{minipage}%
		}%
		\subfigure[128 bits]{
			\begin{minipage}[t]{0.28\linewidth}
				\centering
				\includegraphics[width=\linewidth]{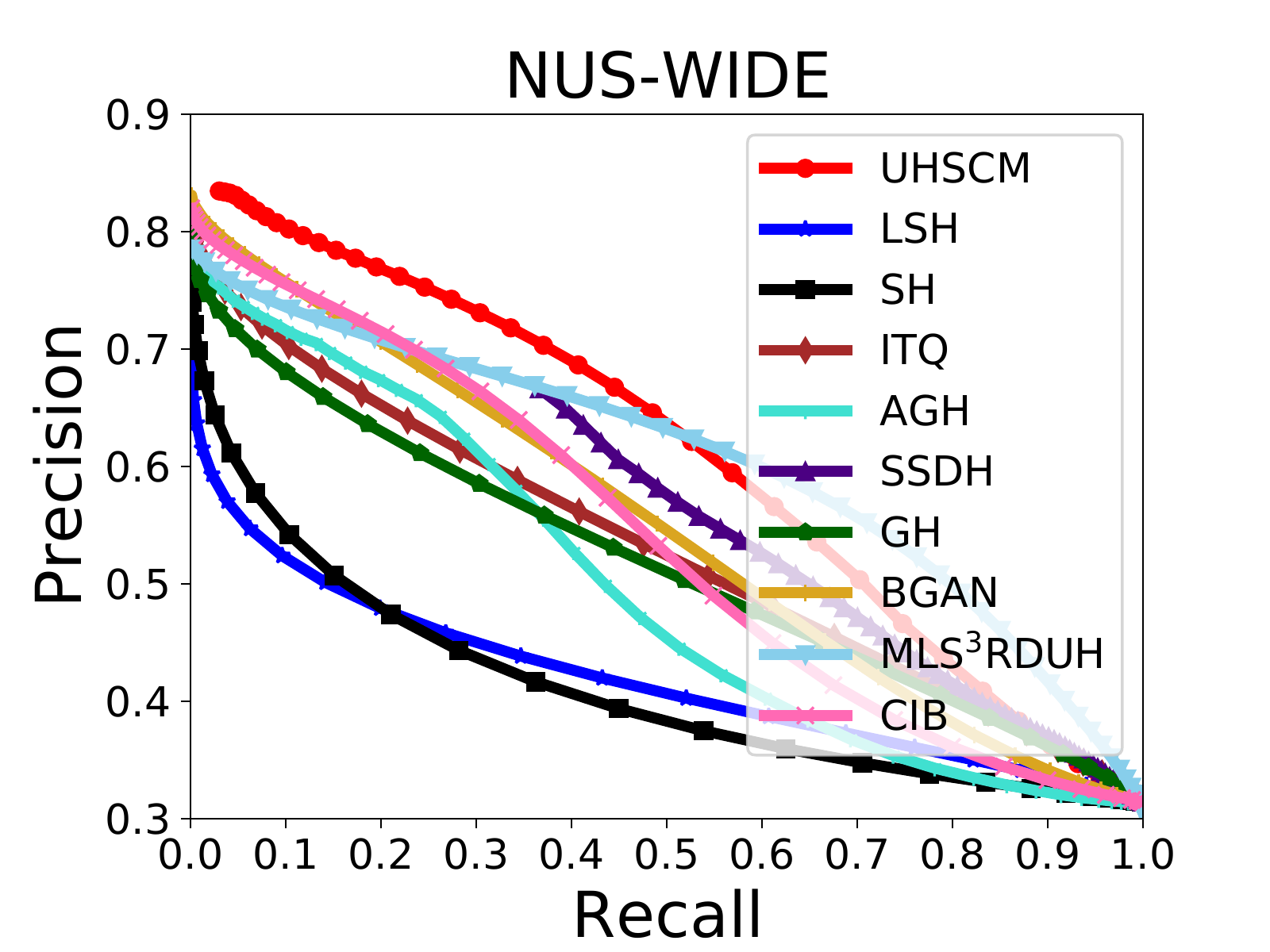}
				%\caption{fig1}
			\end{minipage}%
		}%
		\subfigure[64 bits]{
			\begin{minipage}[t]{0.28\linewidth}
				\centering
				\includegraphics[width=\linewidth]{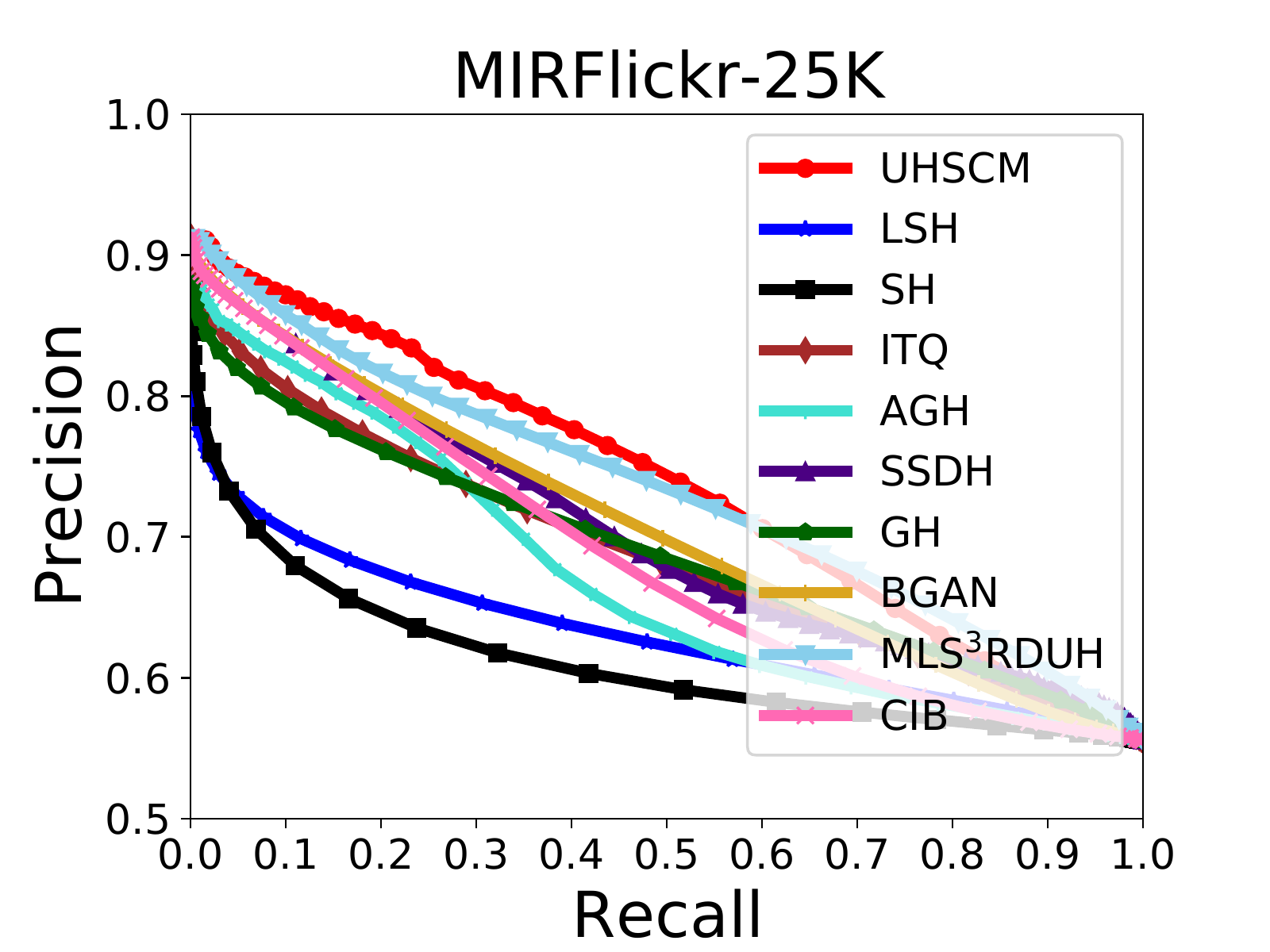}
				%\caption{fig1}
			\end{minipage}%
		}%
		\quad
		\subfigure[128 bits]{
			\begin{minipage}[t]{0.28\linewidth}
				\centering
				\includegraphics[width=\linewidth]{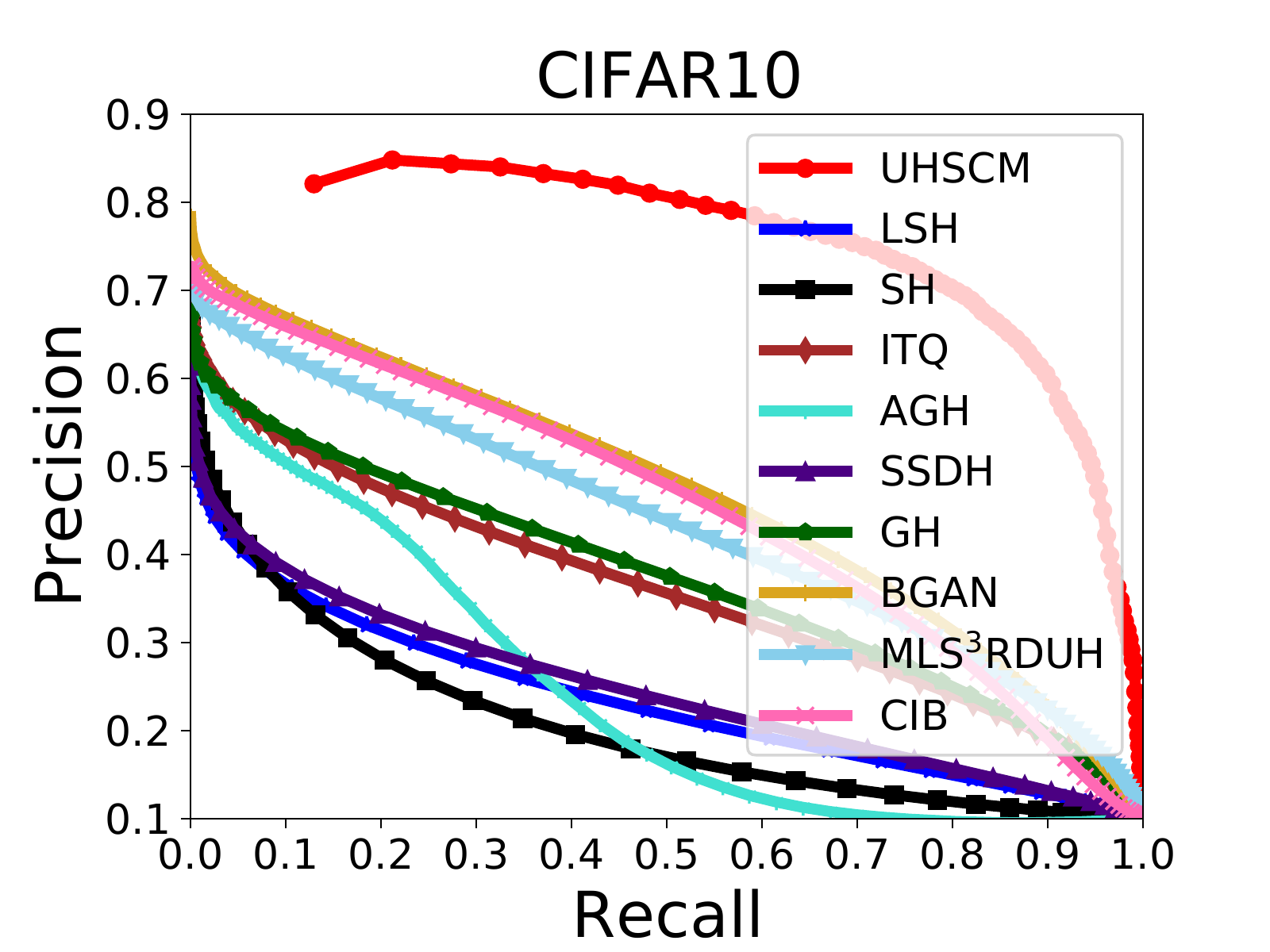}
				%\caption{fig1}
			\end{minipage}%
		}%
		\subfigure[64 bits]{
			\begin{minipage}[t]{0.28\linewidth}
				\centering
				\includegraphics[width=\linewidth]{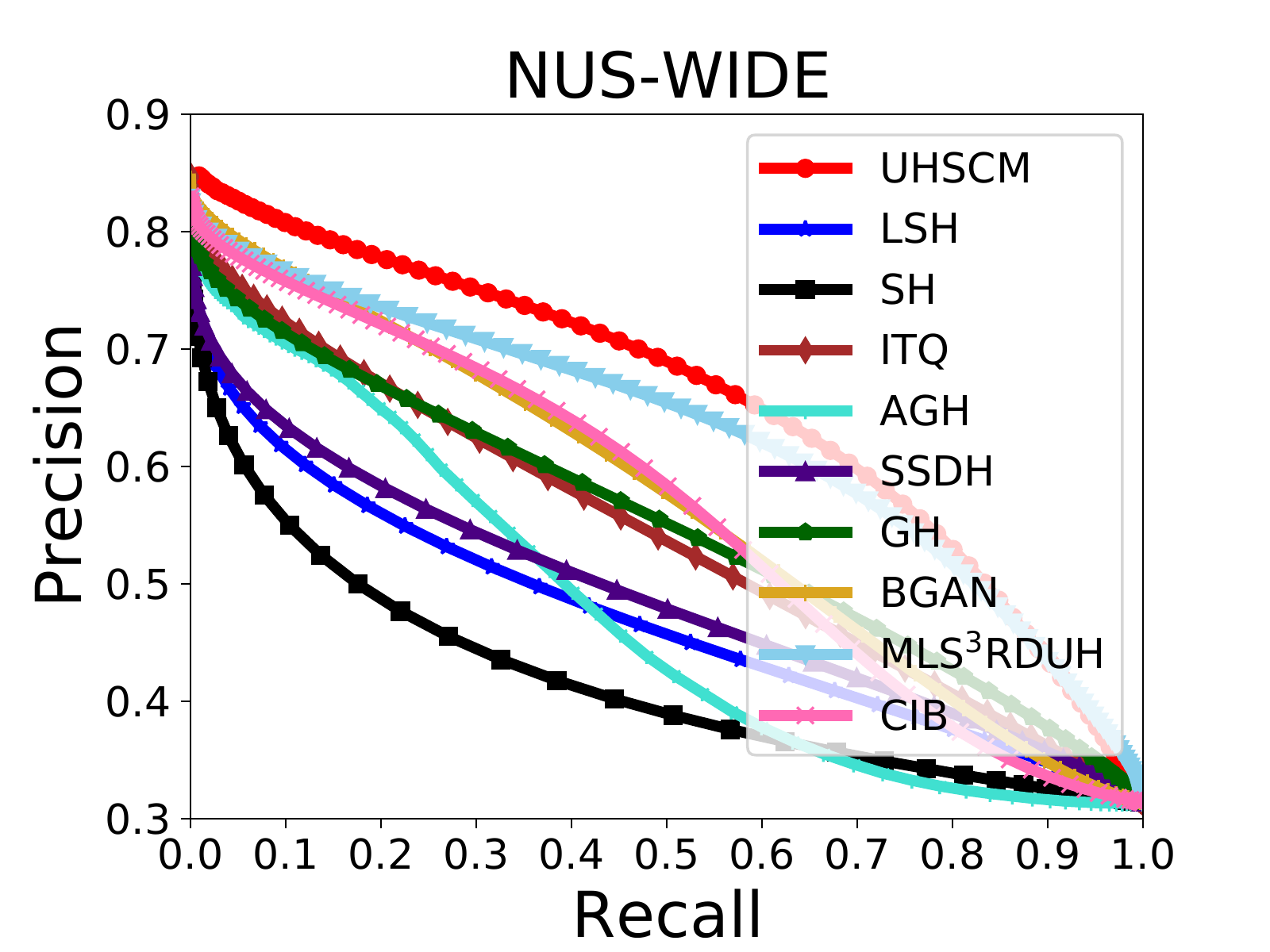}
				%\caption{fig1}
			\end{minipage}%
		}%
		\subfigure[128 bits]{
			\begin{minipage}[t]{0.28\linewidth}
				\centering
				\includegraphics[width=\linewidth]{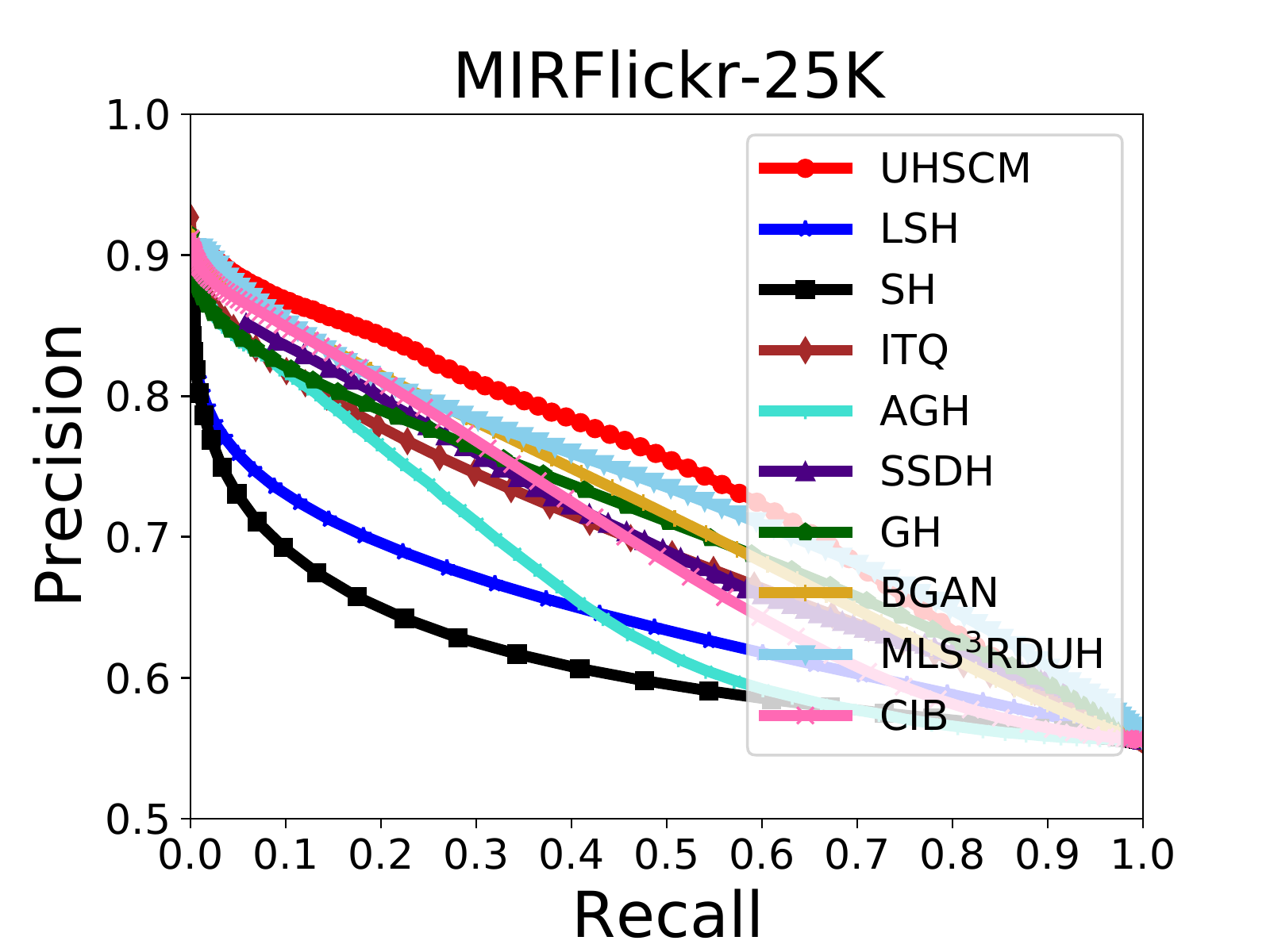}
				%\caption{fig1}
			\end{minipage}%
		}%
		\caption{Precision-recall curves on the three datasets}
		\label{fig_pr}
	\end{figure*} 

	\begin{table*}[]
		\centering
		\begin{tabular}{c|c|cccc|cccc|cccc}
			\toprule[1.2pt]
		\multirow{2}{*}{}  &\multirow{2}{*}{Method} & \multicolumn{4}{c|}{CIFAR10}      & \multicolumn{4}{c|}{NUS-WIDE}  & \multicolumn{4}{c}{MIRFlickr-25K}        \\ \cline{3-14} 
			&&32 bits & 64 bits&96 bits & 128 bits &32 bits & 64 bits&96 bits & 128 bits &32 bits & 64 bits&96 bits & 128 bits  \\ \hline
		
			1	&	UHSCM$_{coco}$ &0.860	&0.866	&0.863	&0.864	&0.771	&0.785	&0.787	&0.786	&0.801	&0.809	&0.810	&0.812 \\ 
				2	&UHSCM$_{nus\&coco}$ &0.858	&0.865	&0.862	&0.857	&0.789	&0.805	&0.809	&0.810	&0.818	&0.824	&0.825	&0.826  \\  \hline \hline
			  3&UHSCM$_{IF}$&0.761	&0.776	&0.779	&0.763	&0.770	&0.795	&0.803	&0.802	&0.772	&0.792	&0.793	&0.789 \\ \hline \hline
				4 &	UHSCM$_{P1}$ &0.823	&0.841	&0.843	&0.849	&0.779	&0.798	&0.805	&0.801	&0.801	&0.815	&0.815 &0.814 \\
		5	&UHSCM$_{P2}$ &0.829	&0.846	&0.847	&0.845	&0.770	&0.789	&0.798	&0.803	&0.793	&0.800	&0.799	&0.800  \\
		6	 &UHSCM$_{avg}$ &0.834	&0.851	&0.850	&0.857	&0.787	&0.805	&0.803	&0.808	&0.810	&0.824	&0.825	&0.826  \\  \hline \hline
			7&UHSCM$_{w/o\ de}$ &0.791	&0.780	&0.798	&0.791	&0.789	&0.805	&0.809	&0.810	&0.818	&0.827	&0.827	&0.825  \\ 
			8&UHSCM$_{c20}$ &0.438	&0.456	&0.469	&0.477	&0.743	&0.764	&0.778	&0.775	&0.769	&0.773	&0.774	&0.777 \\
			9 &UHSCM$_{c30}$&0.515	&0.543	&0.537	&0.550	&0.753	&0.766	&0.779	&0.789	&0.787	&0.792	&0.797	&0.796    \\ 
			10 &UHSCM$_{c40}$&0.601	&0.620	&0.622	&0.637	&0.783	&0.803	&0.806	&0.809	&0.796	&0.798	&0.802	&0.799 \\
			11 &UHSCM$_{c50}$ &0.674	&0.691	&0.685	&0.685	&0.761	&0.781	&0.791	&0.800	&0.810	&0.817	&0.816	&0.815  \\
			12 &UHSCM$_{c60}$ &0.695	&0.697	&0.696	&0.702	&0.771	&0.780	&0.788	&0.792	&0.802	&0.806	&0.807	&0.801 \\
			 \hline \hline
		13 &UHSCM$_{w/o\ MCL}$ &0.730	&0.715	&0.700	&0.680	&0.791	&0.801	&0.800	&0.793	&0.813	&0.819	&0.816	&0.814 \\
			14 &UHSCM$_{CL}$&0.765	&0.800	&0.813	&0.821	&0.783	&0.801	&0.808	&0.810	&0.808	&0.826	&0.830	&0.832           \\ \hline \hline
		 	Ours
			&UHSCM        &0.831&0.850	&0.857&0.853	&0.796	&0.810	&0.813 &0.815	&0.827 &0.834	&0.835	&0.834 \\  \toprule[1.2pt]
		\end{tabular}
		\caption{MAPs of UHSCM and its variants for different numbers of hash bits on the three image datasets.}
		\label{var}
	\end{table*}

\iffalse
\begin{table*}[]
\centering
\begin{tabular}{c|cccc|cccc|cccc}
	\toprule[1.2pt]
	\multirow{2}{*}{Method} & \multicolumn{4}{c|}{CIFAR10}      & \multicolumn{4}{c|}{NUS-WIDE}  & \multicolumn{4}{c}{MIRFlickr-25K}        \\ \cline{2-13} 
	&32 bits & 64 bits&96 bits & 128 bits &32 bits & 64 bits&96 bits & 128 bits &32 bits & 64 bits&96 bits & 128 bits  \\ \hline
	UHSCM$_{1}$ &0.730	&0.715	&0.700	&0.680	&0.791	&0.801	&0.800	&0.793	&0.813	&0.819	&0.816	&0.814 \\
	UHSCM$_{2}$&0.765	&0.800	&0.813	&0.821	&0.783	&0.801	&0.808	&0.810	&0.808	&0.826	&0.830	&0.832           \\ 
	UHSCM$_{3}$&0.761	&0.776	&0.779	&0.763	&0.770	&0.795	&0.803	&0.802	&0.772	&0.792	&0.793	&0.789 \\
	UHSCM$_{4}$ &0.791	&0.780	&0.798	&0.791	&0.789	&0.805	&0.809	&0.810	&0.818	&0.827	&0.827	&0.825  \\ \hline \hline
	UHSCM                &0.831&0.850	&0.857&0.853	&0.796	&0.810	&0.813 &0.815	&0.827 &0.834	&0.835	&0.834 \\  \toprule[1.2pt]
\end{tabular}
\caption{MAP of UHSCM and its variants for different number of bits on the three image datasets.}
\label{var1}
\end{table*}
\fi
	
	\subsubsection{Hash Lookup Protocol}
	When considering the hash lookup protocol, we compute the PR curve for the returned points given any Hamming radius. The PR curve can be obtained by varying the Hamming radius from 0 to $k$ with a step size of 1. The PR curves on 64 and 128bits over the three datasets are shown in Figure \ref{fig_pr}. It can be found that UHSCM outperforms all the state-of-the-art baselines over all three datasets. For example, as shown in Figure \ref{fig_pr} (a) and (d), over the CIFAR10 dataset, the precisions of UHSCM are greatly higher than all the ones of baselines at different recall. Moreover, for the experiments over the NUS-WIDE and MIRFlickr-25K datasets, as shown in Figure \ref{fig_pr} (b), (c), (e) and (f), the PR curves of UHSCM are higher than the ones of baselines on the whole. These results demonstrate that the proposed UHSCM can generate hash codes for similar datapoints in a smaller Hamming radius, i.e., the hash codes generated by the UHSCM are more distinguished than the ones generated by the baselines.
	
	\subsection{Ablation Study}	
	To demonstrate the contributions of different components in our proposed model, we conduct several ablation studies on all the three datasets and the results are shown in Table \ref{var}. Specifically, the UHSCM in the `Ours' row is the final version used in other experiments. It uses the 81 categories of NUS-WIDE as the original concepts and then uses the semantic concept denoising way proposed in Subsection (\ref{scd}) to clean up the concepts and generate a semantic similarity matrix, and finally learns the hashing network by minimizing the Formula (\ref{of}). 
	\subsubsection{\textbf{Semantic Concept Collection}}
	To investigate the effect of the randomly collected concept set, we design  two variants of UHSCM: (1) UHSCM$_{coco}$ uses all  80 categories of MS COCO \cite{lin2014microsoft} as the original concepts and with the other setting same as that of UHSCM. (2) UHSCM$_{nus\&coco}$ combines all the categories of NUS-WIDE and those of MS COCO and then obtains a total of 153 different categories as the original concepts.
	
	Based on these results shown in `1' and `2' rows of Table \ref{var}, it can be observed: (1) Compared with the UHSCM$_{coco}$, our proposed UHSCM using the 81 categories of NUS-WIDE as the original concepts can achieve better retrieval performance on the NUS-WIDE and MIRFlickr-25K datasets but UHSCM$_{coco}$ performs better on the CIFAR10 dataset. This may be because the 80 categories of MS COCO contain most of the classes of CIFAR10, but are very different with the 21 classes of NUS-WIDE and the 24 classes of MIRFlickr-25K used in our experimental settings. Moreover, the 81 categories of NUS-WIDE contain most of the 21 classes of NUS-WIDE and the 24 classes of MIRFlickr-25K. Hence, UHSCM$_{coco}$ achieves better performance on CIFAR10 dataset and UHSCM  performs  better on the NUS-WIDE and MIRFlickr-25K datasets. These results show that for two randomly selected concept sets whose number of concepts is almost equal, our proposed method can achieve better retrieval performance when using the concept set with more concepts related to the classes of the experimental dataset as the original concept set. (2) Comparing the UHSCM$_{coco}$, UHSCM$_{nus\&coco}$ and UHSCM, UHSCM$_{coco}$ achieves the best performance on the CIFAR10 dataset, and UHSCM achieves the best performance on the NUS-WIDE and MIRFlickr-25K datasets. Although the 153 categories collected from MS COCO and NUS-WIDE contain more concepts related to the classes of the three experimental datasets, the UHSCM$_{nus\&coco}$ still cannot achieves the best performance on the three datasets. This may be because the 153 categories  contain more useless concepts that confuse the CLIP model and then reduce the quality of the defined semantic similarity matrix. Hence, these results reveal that when lots of useless concepts are contained in the original collected concept set, they will harm the retrieval performance of our hashing model, which also means that it is necessary to denoise the concept set.

	\begin{figure*}[t]
		\centering
		\subfigure[]{
			\begin{minipage}[t]{0.2\linewidth}
				\centering
				\includegraphics[width=\linewidth]{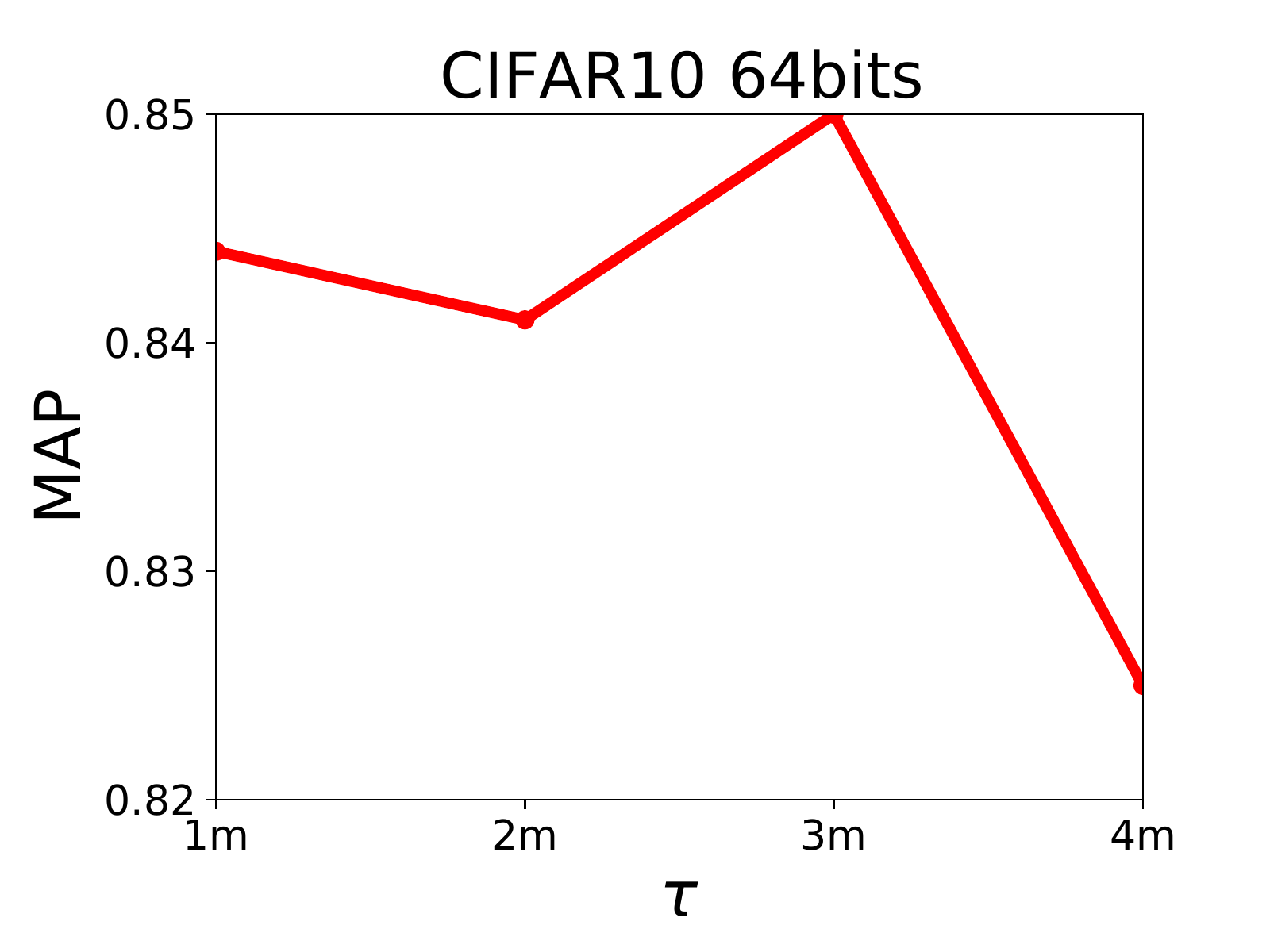}
				%\caption{fig1}
			\end{minipage}%
		}%
		\subfigure[]{
			\begin{minipage}[t]{0.2\linewidth}
				\centering
				\includegraphics[width=\linewidth]{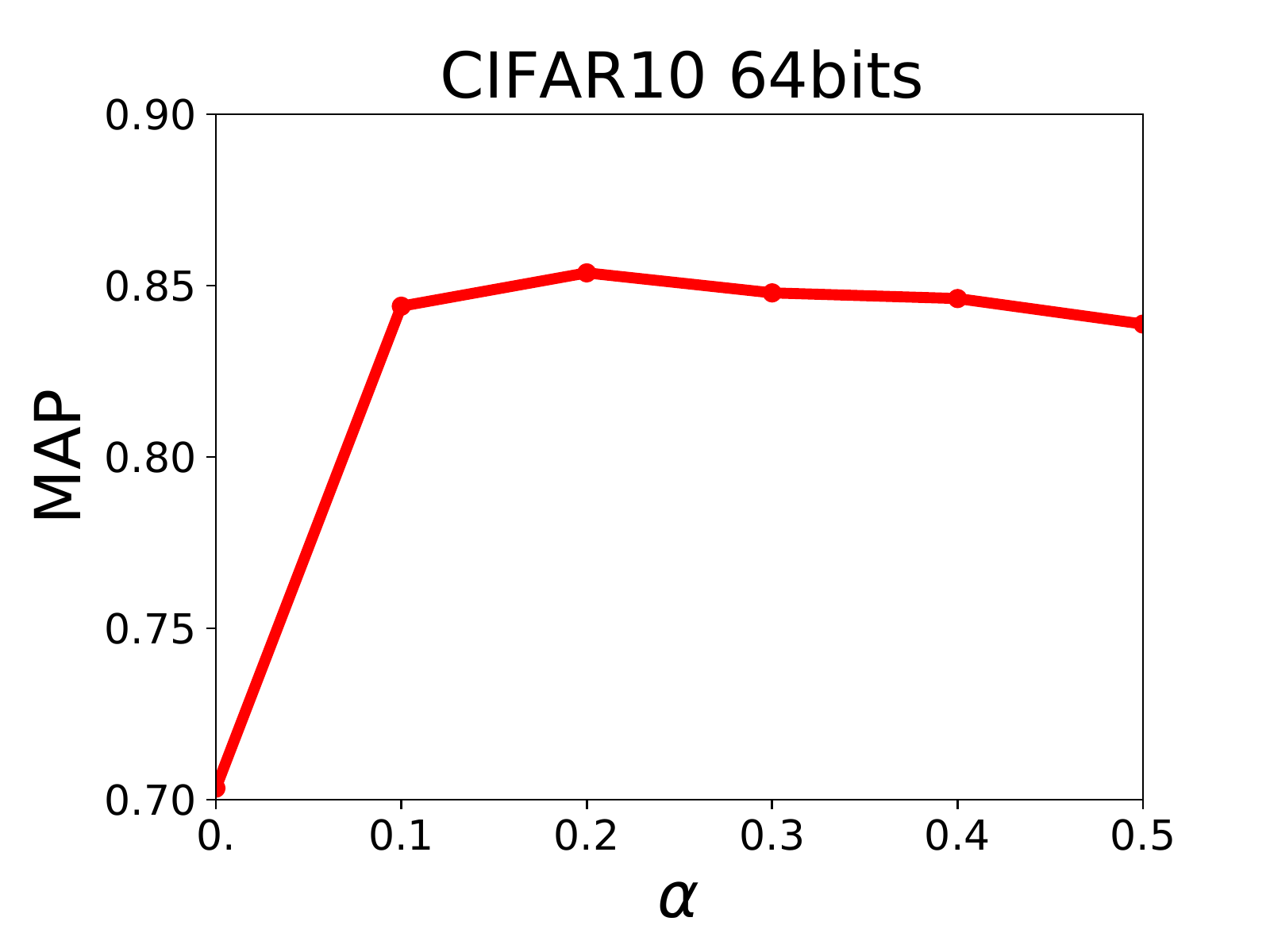}
				%\caption{fig1}
			\end{minipage}%
		}%
		\subfigure[]{
			\begin{minipage}[t]{0.2\linewidth}
				\centering
				\includegraphics[width=\linewidth]{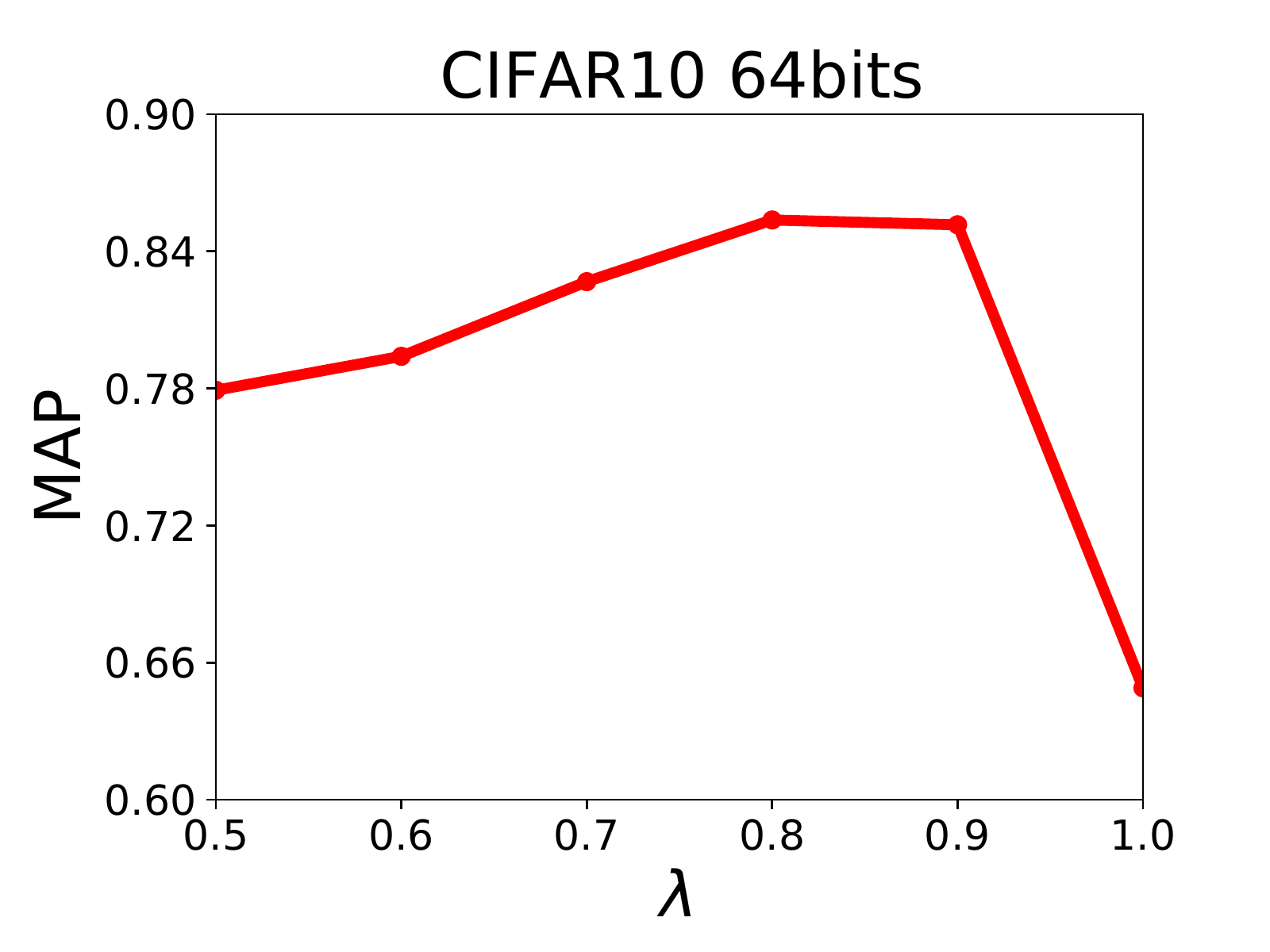}
				%\caption{fig1}
			\end{minipage}%
		}%
		\subfigure[]{
			\begin{minipage}[t]{0.2\linewidth}
				\centering
				\includegraphics[width=\linewidth]{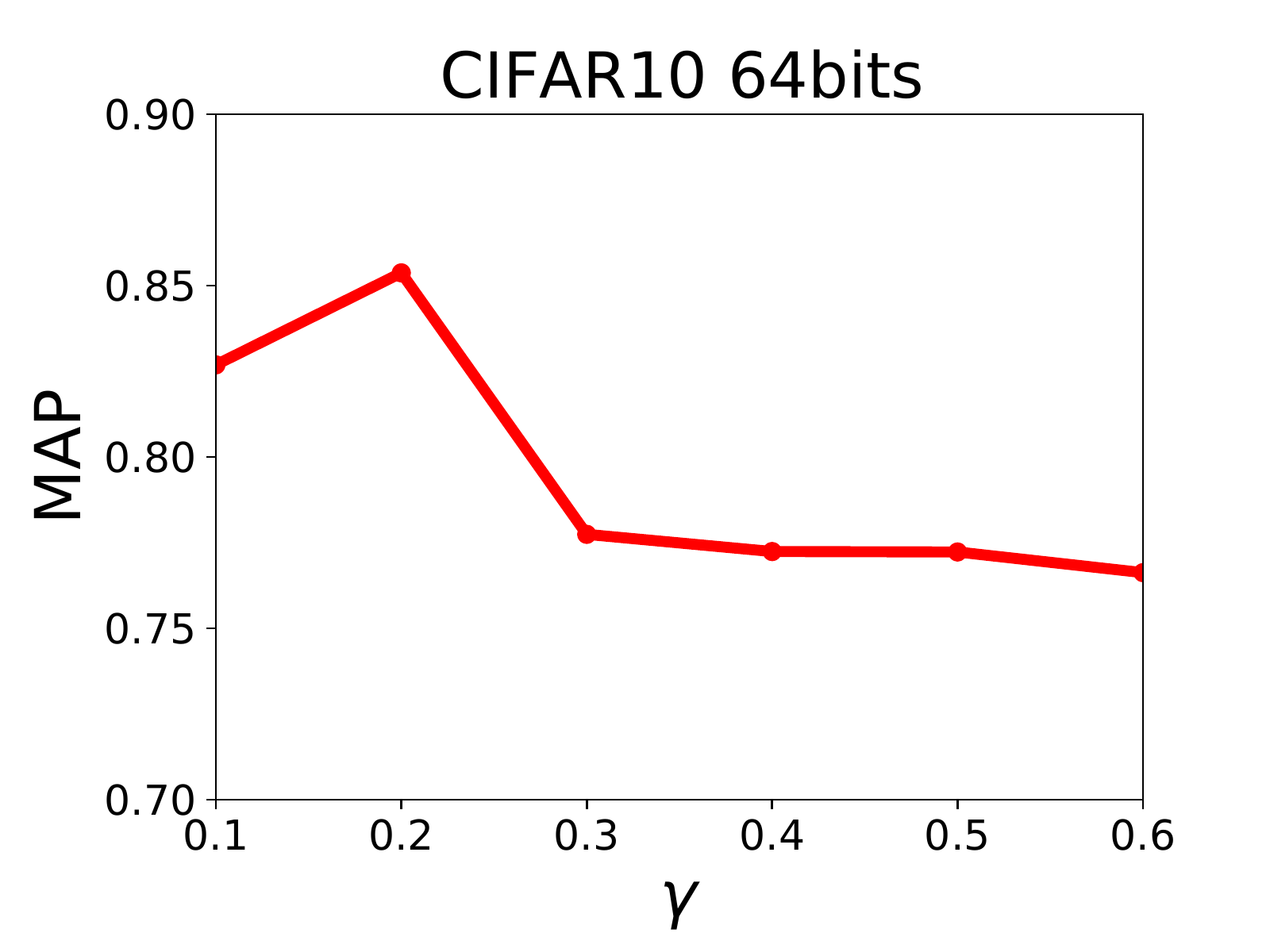}
				%\caption{fig1}
			\end{minipage}%
		}%
		\subfigure[]{
			\begin{minipage}[t]{0.2\linewidth}
				\centering
				\includegraphics[width=\linewidth]{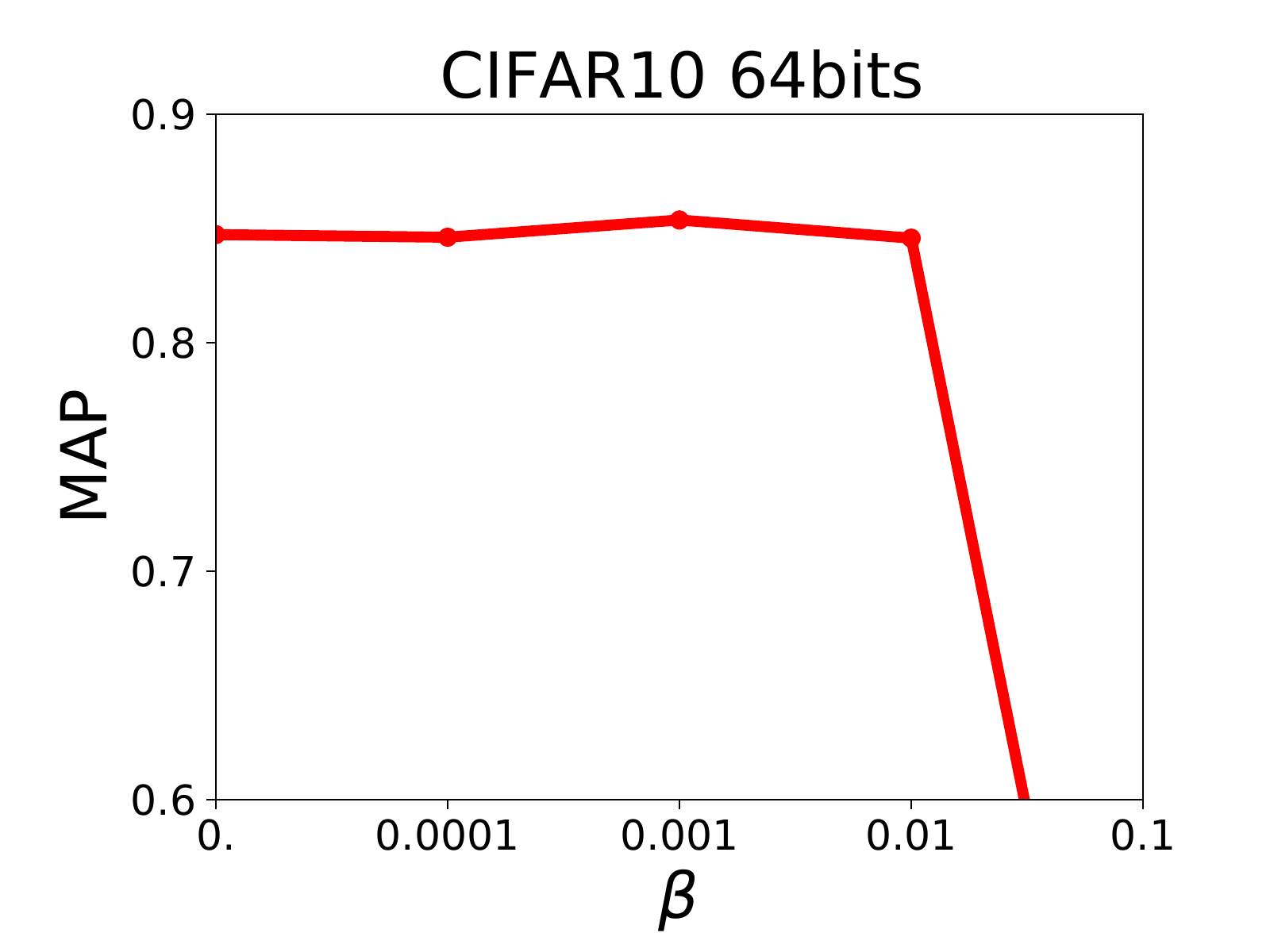}
				%\caption{fig1}
			\end{minipage}%
		}%
		\quad
		\subfigure[]{
			\begin{minipage}[t]{0.2\linewidth}
				\centering
				\includegraphics[width=\linewidth]{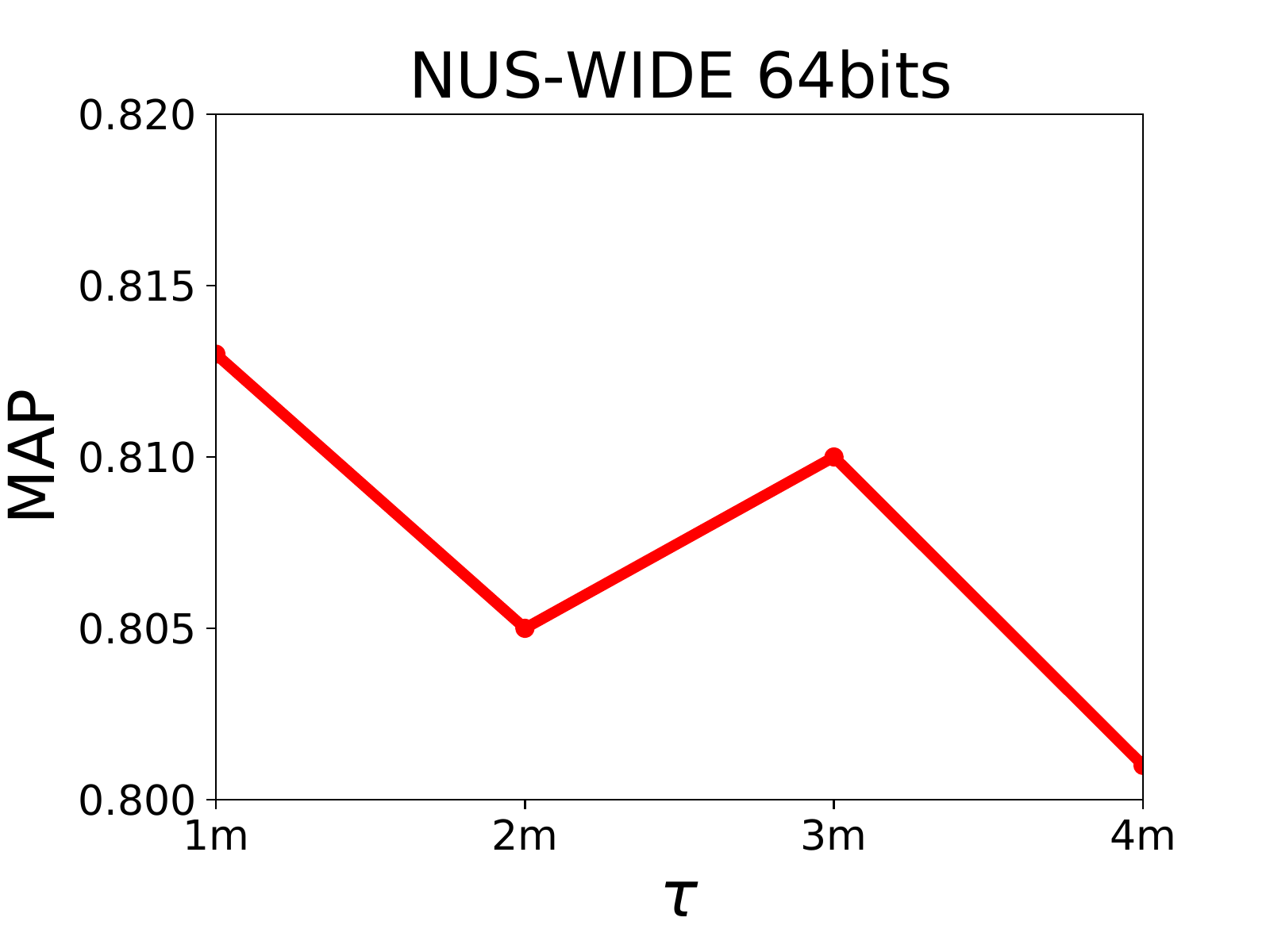}
				%\caption{fig1}
			\end{minipage}%
		}%
		\subfigure[]{
			\begin{minipage}[t]{0.2\linewidth}
				\centering
				\includegraphics[width=\linewidth]{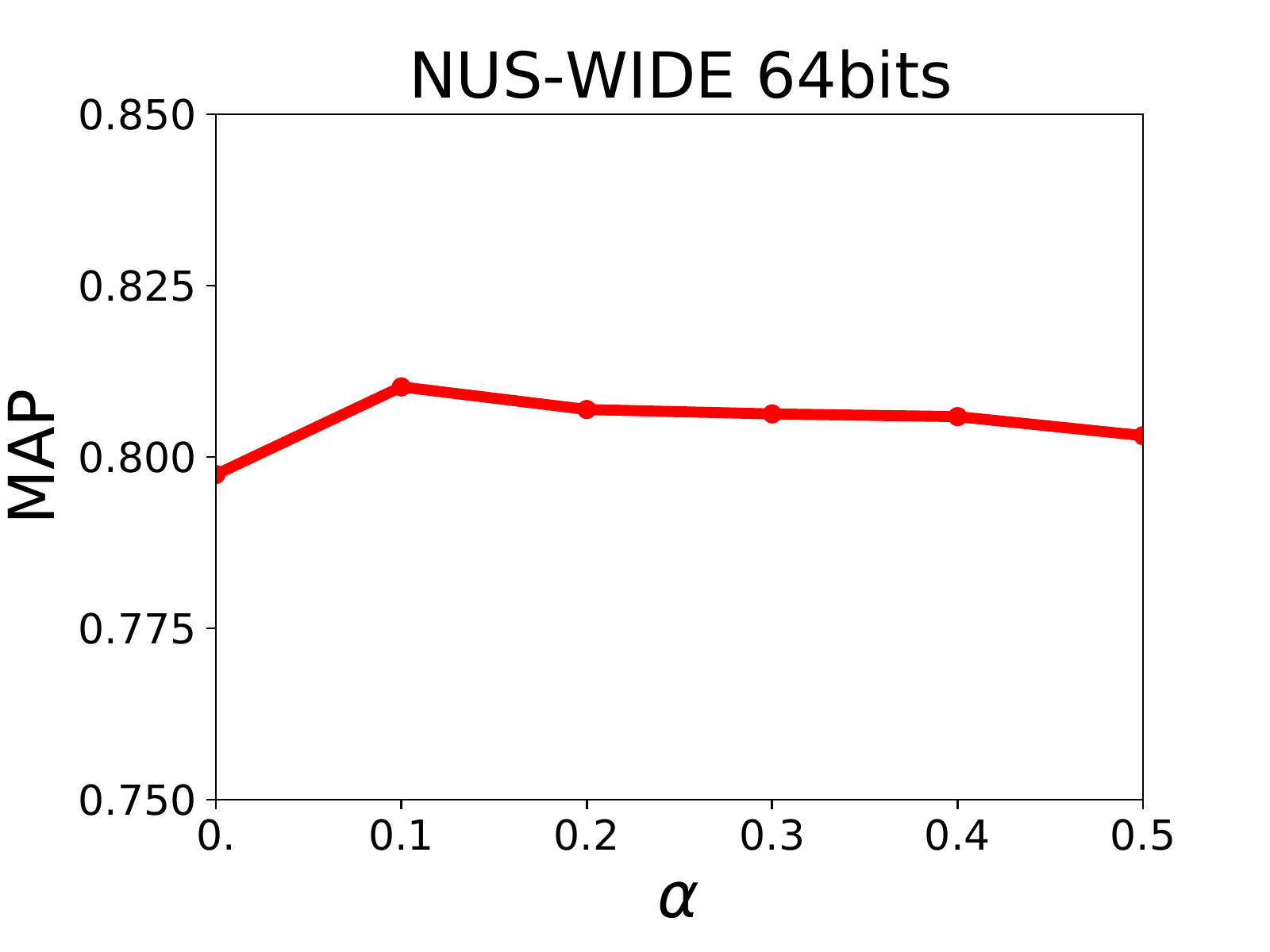}
				%\caption{fig1}
			\end{minipage}%
		}%
		\subfigure[]{
			\begin{minipage}[t]{0.2\linewidth}
				\centering
				\includegraphics[width=\linewidth]{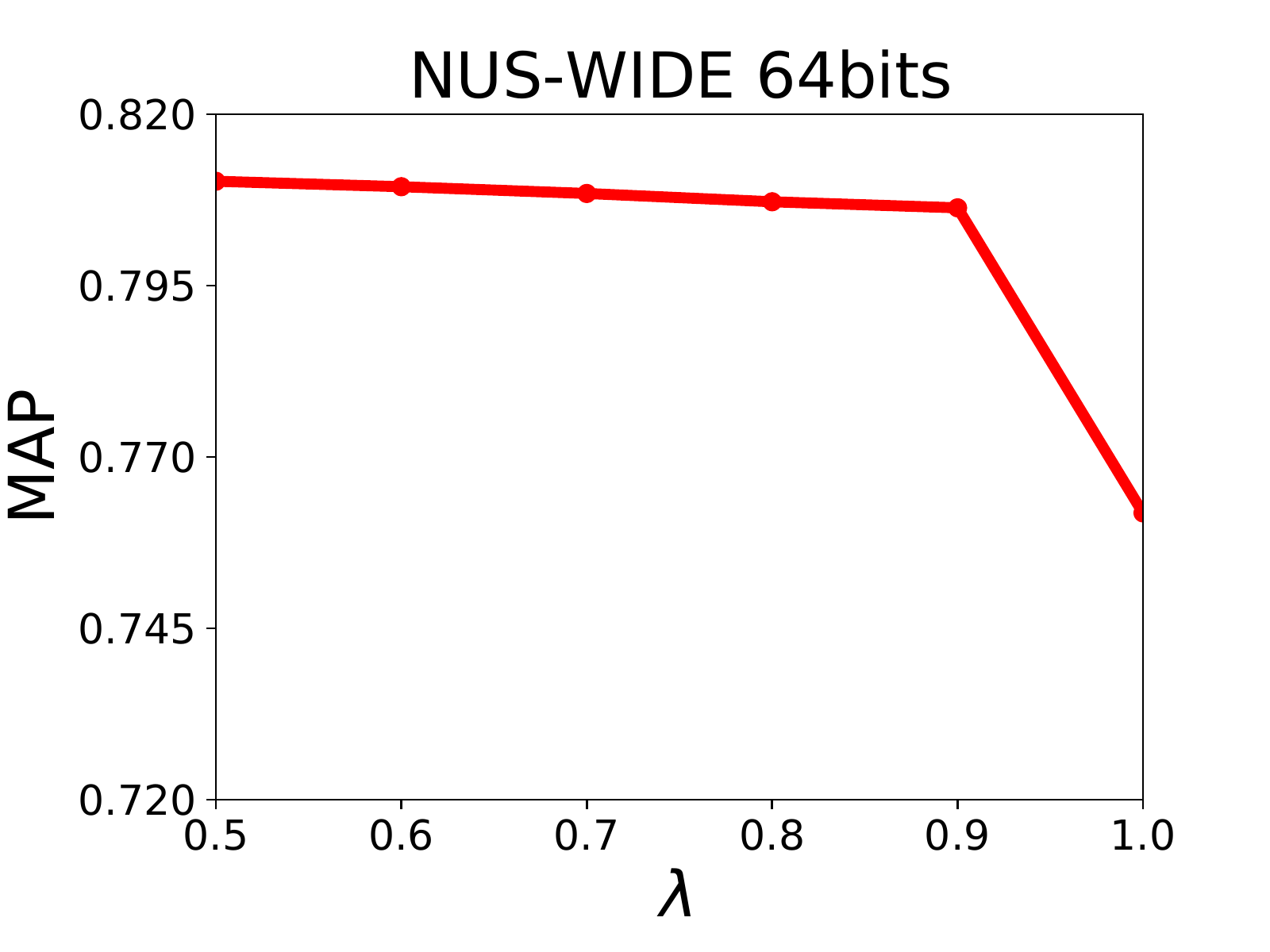}
				%\caption{fig1}
			\end{minipage}%
		}%
		\subfigure[]{
			\begin{minipage}[t]{0.2\linewidth}
				\centering
				\includegraphics[width=\linewidth]{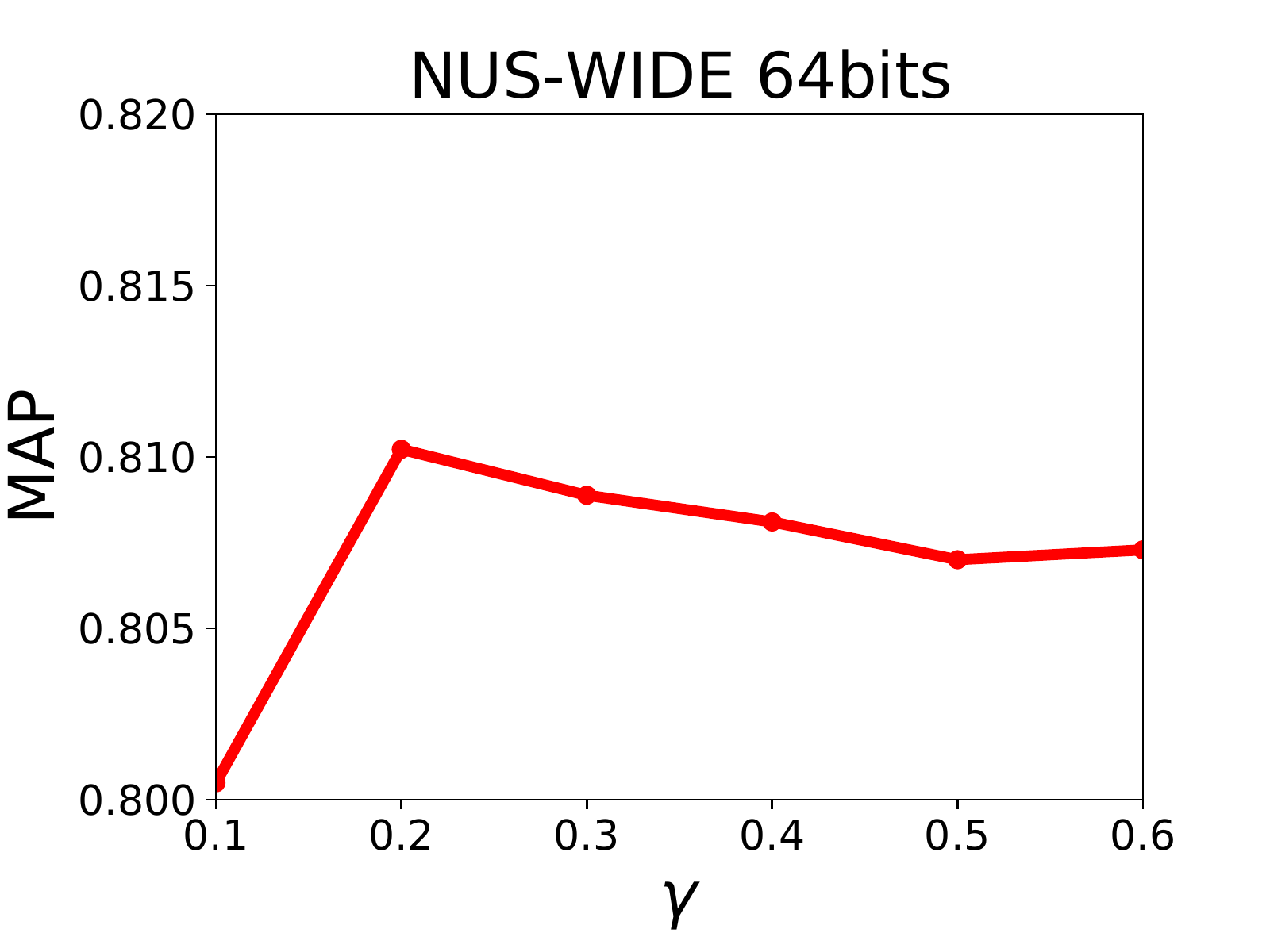}
				%\caption{fig1}
			\end{minipage}%
		}%
		\subfigure[]{
			\begin{minipage}[t]{0.2\linewidth}
				\centering
				\includegraphics[width=\linewidth]{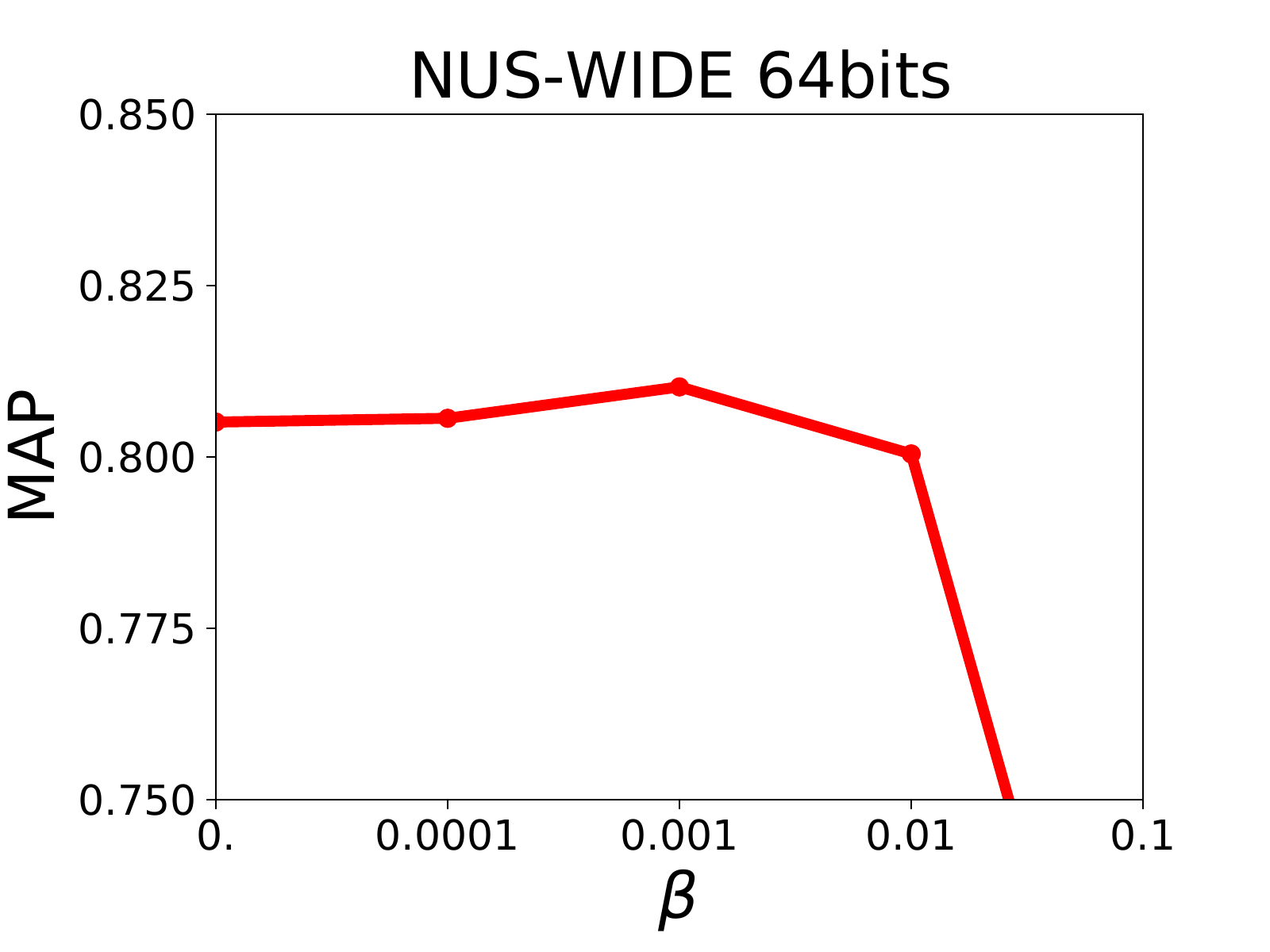}
				%\caption{fig1}
			\end{minipage}%
		}%
		\quad
		\subfigure[]{
			\begin{minipage}[t]{0.2\linewidth}
				\centering
				\includegraphics[width=\linewidth]{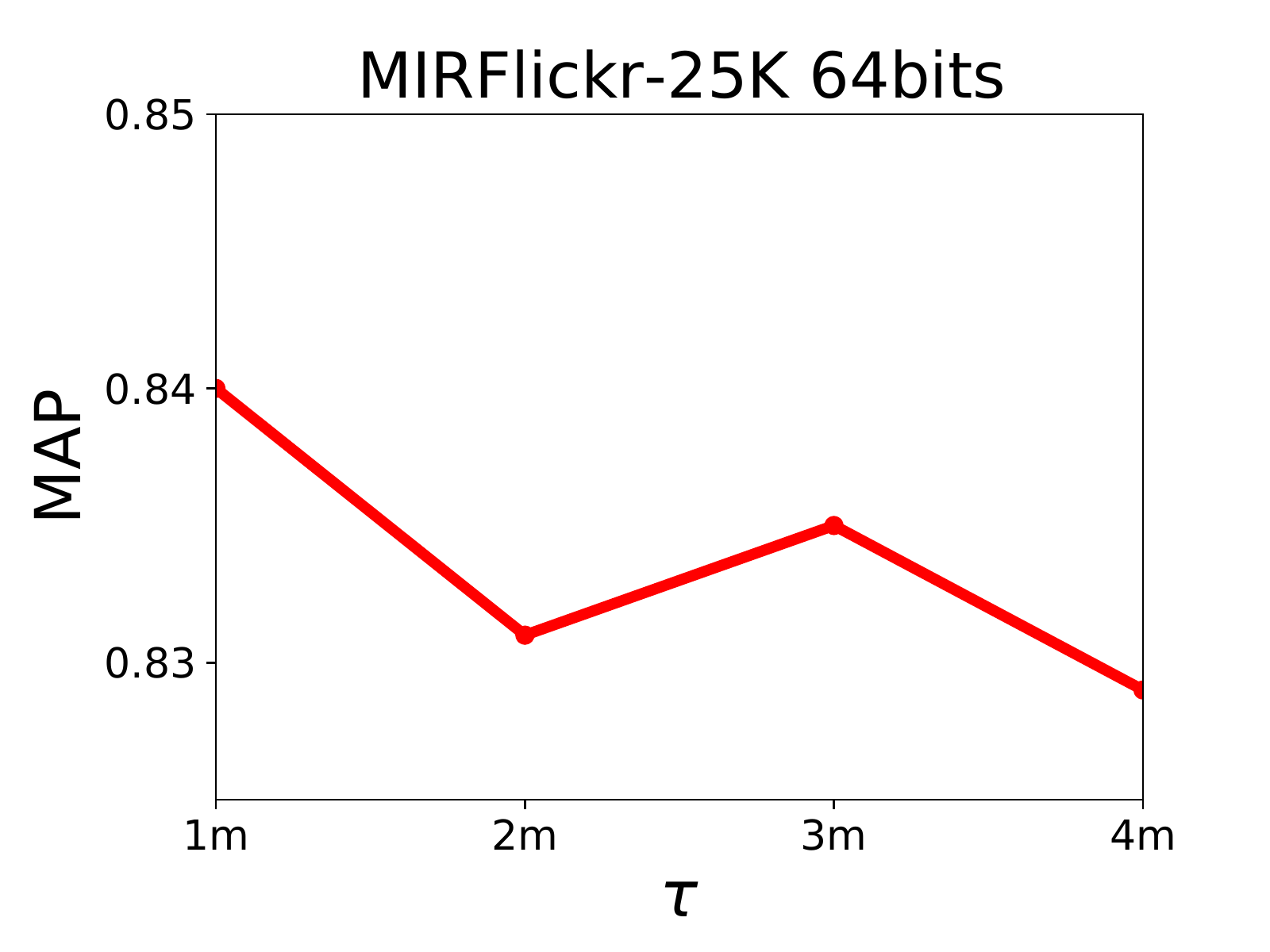}
				%\caption{fig1}
			\end{minipage}%
		}%
		\subfigure[]{
			\begin{minipage}[t]{0.2\linewidth}
				\centering
				\includegraphics[width=\linewidth]{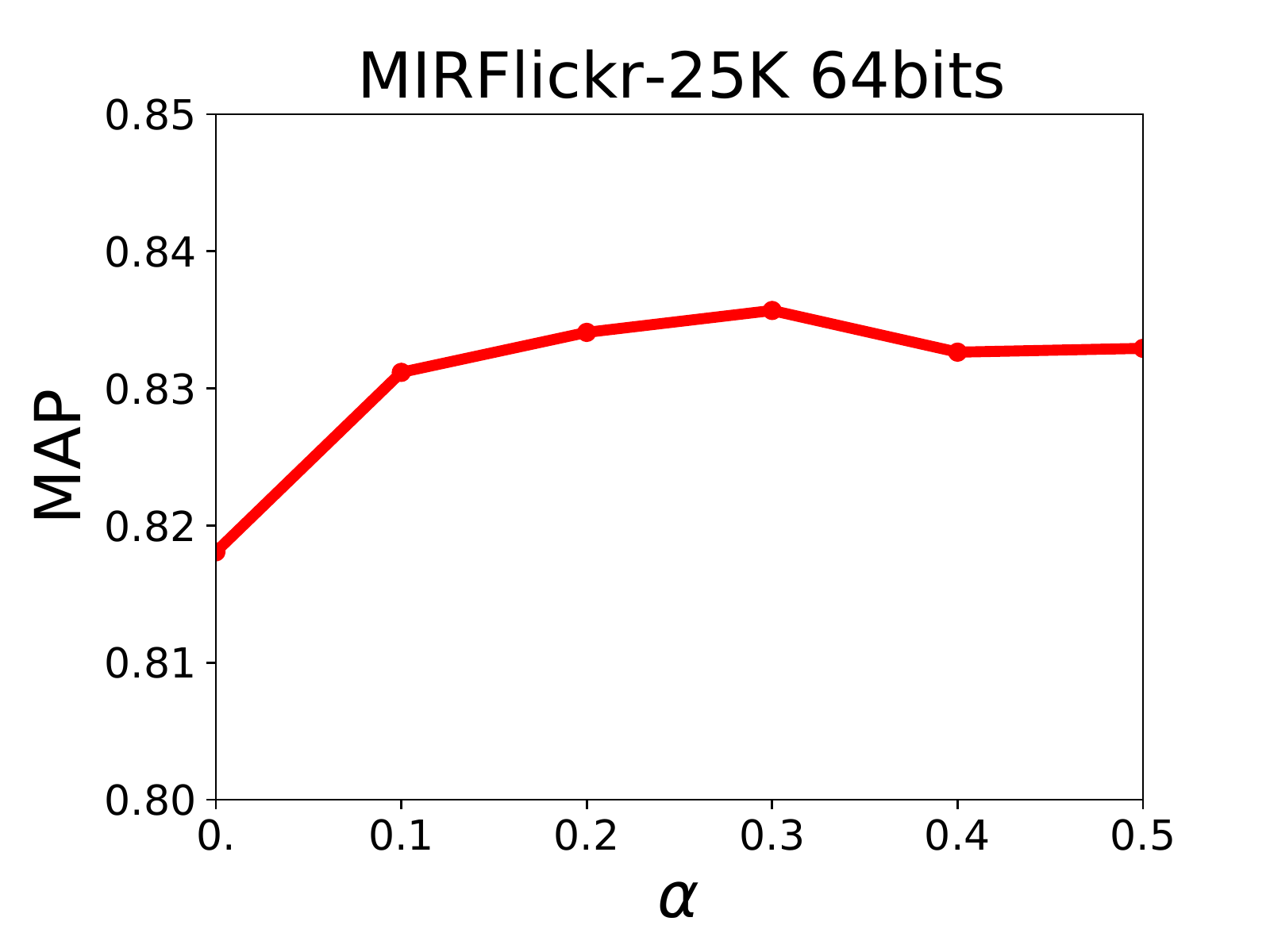}
				%\caption{fig1}
			\end{minipage}%
		}%
		\subfigure[]{
			\begin{minipage}[t]{0.2\linewidth}
				\centering
				\includegraphics[width=\linewidth]{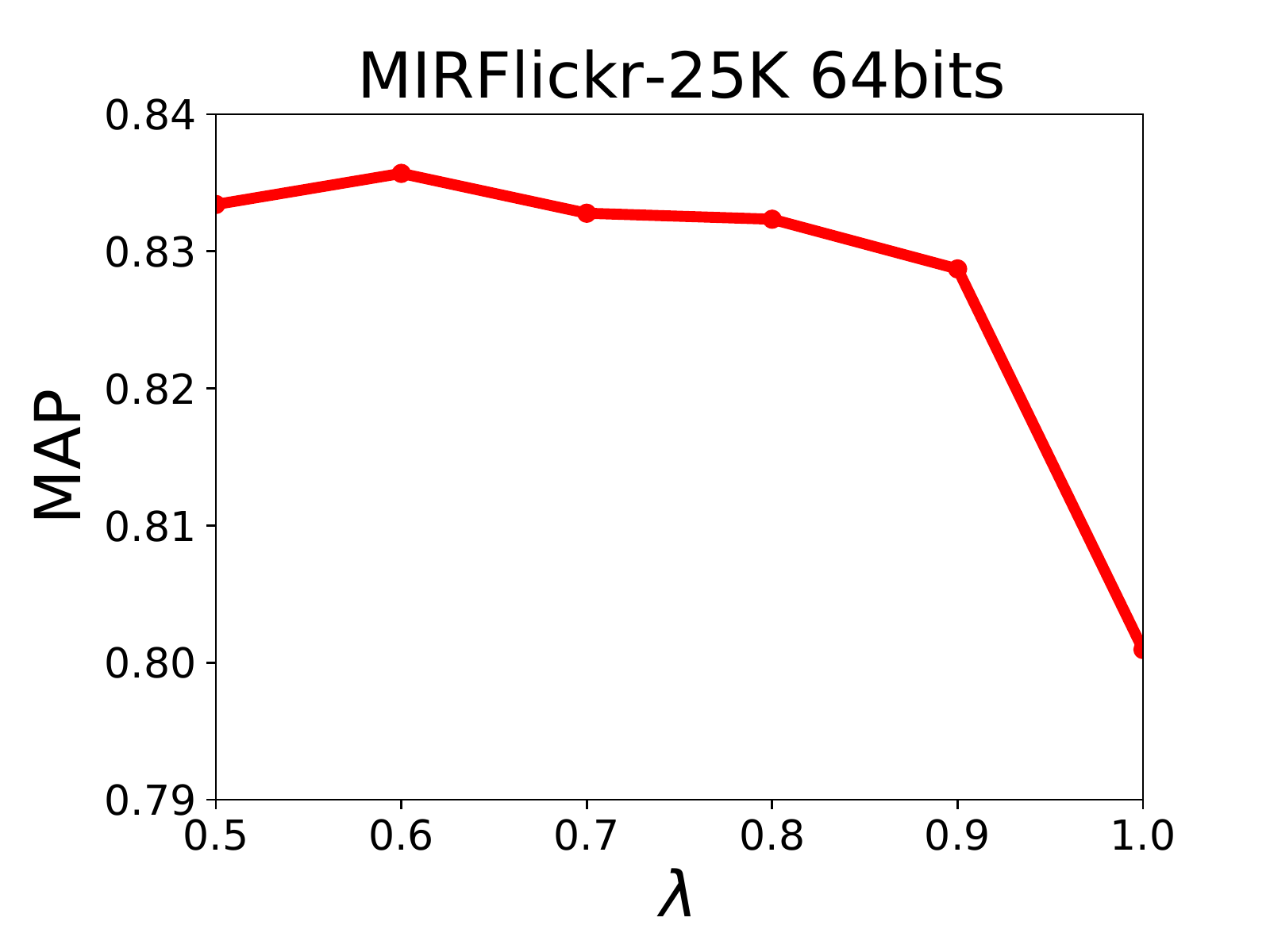}
				%\caption{fig1}
			\end{minipage}%
		}%
		\subfigure[]{
			\begin{minipage}[t]{0.2\linewidth}
				\centering
				\includegraphics[width=\linewidth]{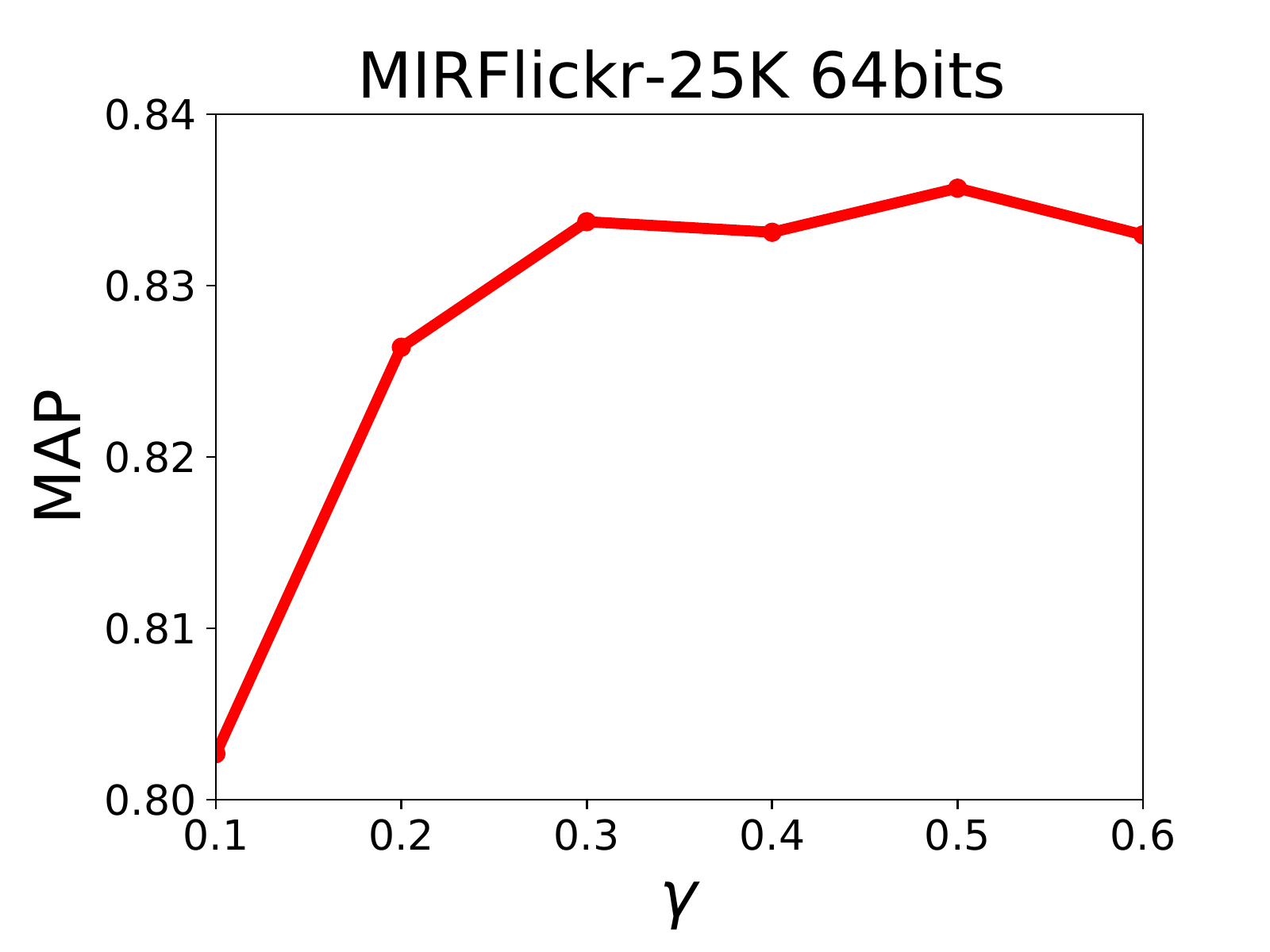}
				%\caption{fig1}
			\end{minipage}%
		}%
		\subfigure[]{
			\begin{minipage}[t]{0.2\linewidth}
				\centering
				\includegraphics[width=\linewidth]{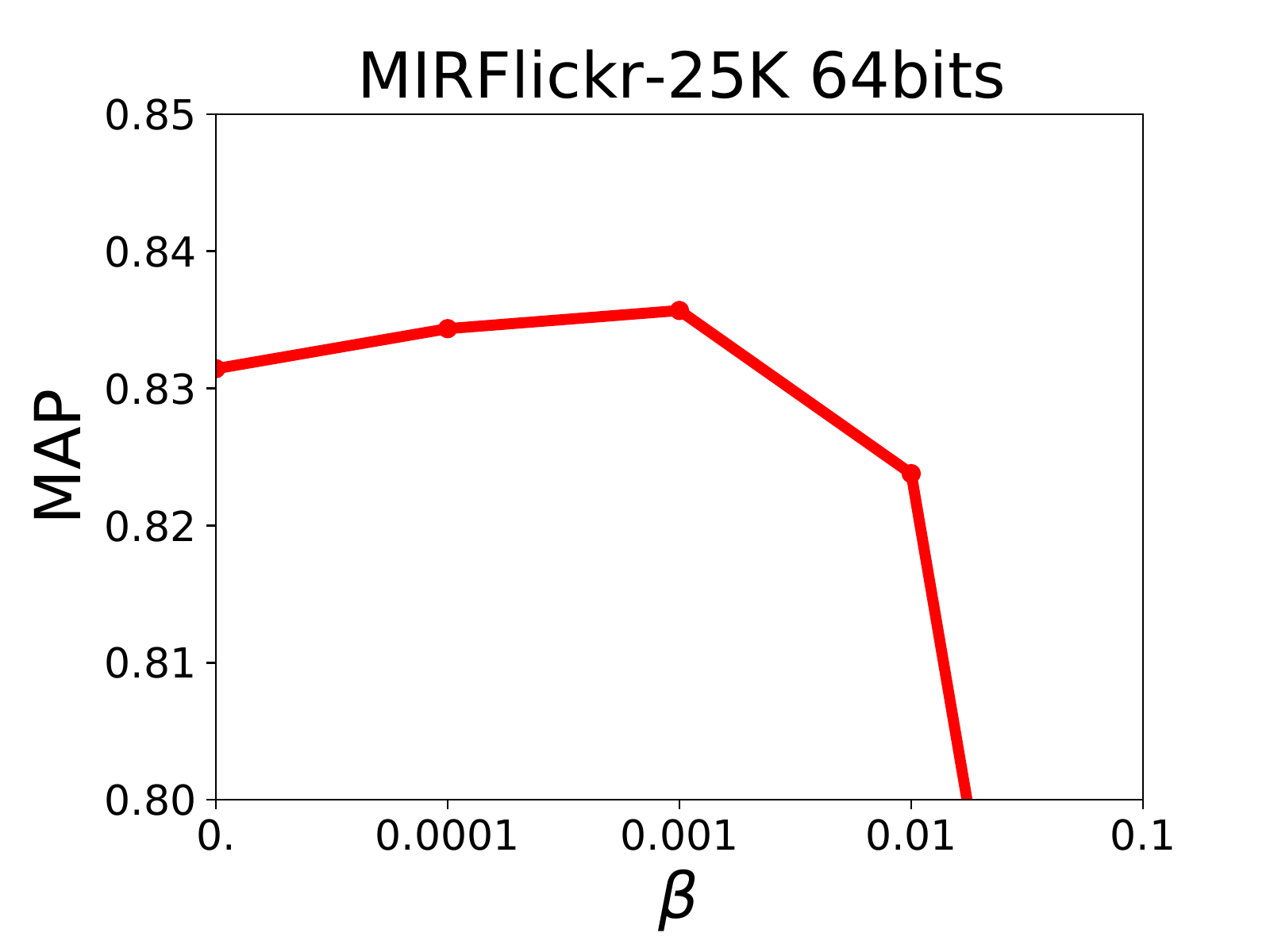}
				%\caption{fig1}
			\end{minipage}%
		}%
		\caption{A sensitivity analysis of the hyper-parameters}
		\label{fig_para}
	\end{figure*}

	\subsubsection{\textbf{Semantic Concept Mining}} We design a variant UHSCM$_{IF}$ which directly uses the image features extracted by the CLIP model to construct the semantic similarity matrix without mining the concept semantic distributions through the prompt engineering. As shown in the `3' row of Table \ref{var}, compared with the UHSCM$_{IF}$, the MAP results of UHSCM achieve an average increase of 7.8\%, 1.6\% and 4.6\% on the CIFAR10, NUS-WIDE and MIRFlickr-25K datasets, respectively. It means that the semantic similarity matrix constructed with the concept information  is  high-quality. These results show that mining the concepts contained in images to construct similarity will improve the retrieval performance of hashing network. 
	
	\subsubsection{\textbf{Prompt Template}} 
	Here, we study the impact of prompt template. Our proposed UHSCM uses "a photo of the {$\boldsymbol{c}_i$}" as the prompt template to construct the corresponding text $\boldsymbol{t}_i$ for {$\boldsymbol{c}_i$}, and the UHSCM$_{P1}$ and UHSCM$_{P2}$ use  "the {$\boldsymbol{c}_i$}" and "it contains the {$\boldsymbol{c}_i$}" as the prompt template, respectively. The UHSCM$_{avg}$ is a variant whose semantic similarity matrix is the mean value of the similarity matrices of UHSCM, UHSCM$_{P1}$ and UHSCM$_{P2}$. As the results shown in the `4' to `6' rows of Table \ref{var}, it can be found that using the template "a photo of the {$\boldsymbol{c}_i$}" will help our hashing model achieve the best performance. These results show that the prompt template also plays a key role in improving the retrieval performance of our hashing model.

	\subsubsection{\textbf{Semantic Concept Denoising}}
	To investigate the effect of our proposed semantic concept denoising component, we propose some variants. First we design a UHSCM$_{w/o\ de}$  whose  semantic similarity matrix is constructed without denoising the set of randomly collected concepts, i.e., calculated by the Formula (\ref{acal}). Next, we define a series of clustering based variants through clustering the original randomly selected concepts into $n$  clusters by K-means \cite{macqueen1967classification} and then using the $n$ clusters as the final concepts to generate the semantic similarity matrix, and these variants are termed as UHSCM$_{cn}$ where $n$ denotes the number of clusters.  Based on the results shown in the `7' to `12' rows of Table \ref{var}, two observations are obtained: (1) The semantic similarity matrix construct with the  denoised concepts is more high-quality than that constructed with the randomly collected concepts.  For example, compared with the UHSCM$_{w/o\ de}$, the MAP results of UHSCM achieve an average increase of 5.8\%, 0.5\% and 0.8\% on the CIFAR10, NUS-WIDE and MIRFlickr-25K datasets, respectively. These results demonstrate that it is necessary to denoise the original randomly selected concept set. (2) Our proposed semantic concept denoising component is more useful than the clustering way. For instance, compared with the best clustering based variant UHSCM$_{c50}$ on MIRFlickr-25K dataset, the MAP results of UHSCM achieve an average increase of 1.8\%.

	\subsubsection{\textbf{Modified Contrastive Loss}}
	To investigate the effect of our proposed modified contrastive loss, we proposed two variants: (1) UHSCM$_{w/o\ MCL}$ constructs the hashing loss function without the modified contrastive loss $\mathcal{L}_c$ as regularization item; UHSCM$_{CL}$ replace the modified contrastive loss $\mathcal{L}_c$ with the original contrastive loss $\mathcal{J}_c$ to construct the hashing loss function. The corresponding results are shown in the `13' and `14' rows of Table \ref{var}. From these results, the following insight can be obtained: (1) With the modified contrastive loss based regularization to optimize the hashing model, it will improve the image retrieval performance. For example, compared with the UHSCM$_{w/o MCL}$, the MAP results of UHSCM achieve an average increase of 14.4\%, 1.2\% and 1.7\% on the CIFAR10, NUS-WIDE and MIRFlickr-25K datasets, respectively. These results demonstrate that the modified contrastive loss $\mathcal{L}_c$ can further take a good use of the constructed semantic similarity information to guide hashing model generate distinguished hash codes. (2) Our modified contrastive loss $\mathcal{L}_c$ is more useful than the original one $\mathcal{J}_c$ for the hashing model.  For example, compared with the UHSCM$_{CL}$, the MAP results of UHSCM achieve an average increase of 4.8\%, 0.8\% and 0.9\% on the CIFAR10, NUS-WIDE and MIRFlickr-25K datasets, respectively.

	\begin{table}[]
	\centering
	\begin{tabular}{c|c|c|c}
		\toprule[1.2pt]
			Method & CIFAR10   & NUS-WIDE  & MIRFlickr-25K       \\ \hline
			SSDH             &24.9	&21.2	&20.8  \\ 
			GH             &25.7	&28.4	&21.3	   \\
			BGAN                   &78.1	&83.3	&66.1    \\ 
			MLS$^3$RDUH           &132.7	&126.5	&114.7	 \\ 
			CIB                     &31.5	&34.6	&18.5    \\ \hline\hline
			UHSCM                &27.3	&35.7	&20.2	  \\  \toprule[1.2pt]
	\end{tabular}
	\caption{The time consumption (in minute) of our method and baselines on the three image datasets.}
	\label{time}
\end{table}

\begin{figure*}[t]
	\centering
	\subfigure[UHSCM]{
		\begin{minipage}[t]{0.25\linewidth}
			\centering
			\includegraphics[width=\linewidth]{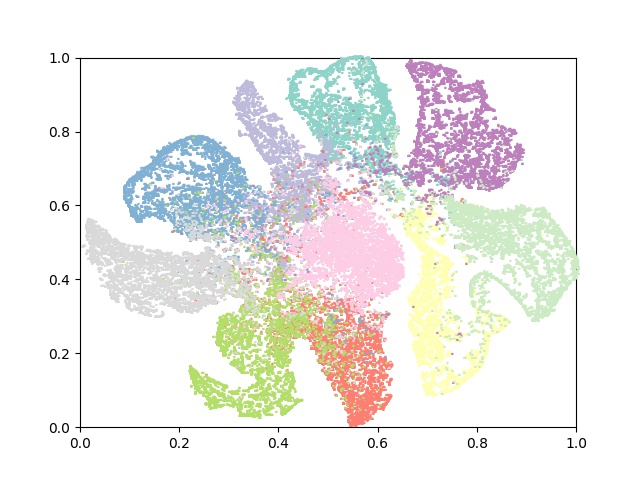}
			%\caption{fig1}
		\end{minipage}%
	}%
	\subfigure[CIB]{
		\begin{minipage}[t]{0.25\linewidth}
			\centering
			\includegraphics[width=\linewidth]{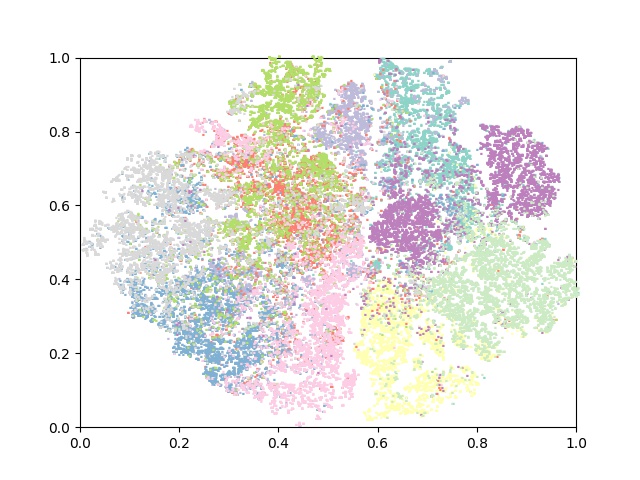}
			%\caption{fig1}
		\end{minipage}%
	}%
	\subfigure[MLS$^3$RDUH]{
		\begin{minipage}[t]{0.25\linewidth}
			\centering
			\includegraphics[width=\linewidth]{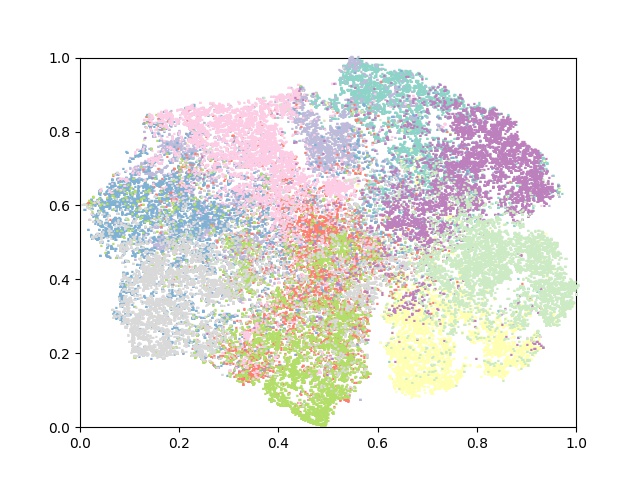}
			%\caption{fig1}
		\end{minipage}%
	}%
	\subfigure[BGAN]{
		\begin{minipage}[t]{0.25\linewidth}
			\centering
			\includegraphics[width=\linewidth]{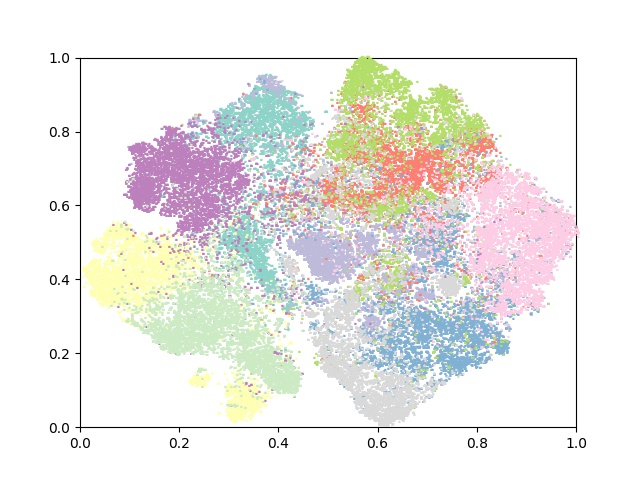}
			%\caption{fig1}
		\end{minipage}%
	}%
	\caption{t-SNE visualization on the CIFAR-10 dataset.}
	\label{fig_tsne}
\end{figure*} 
\begin{figure}[tb]
	\centering
	\includegraphics[width=\linewidth]{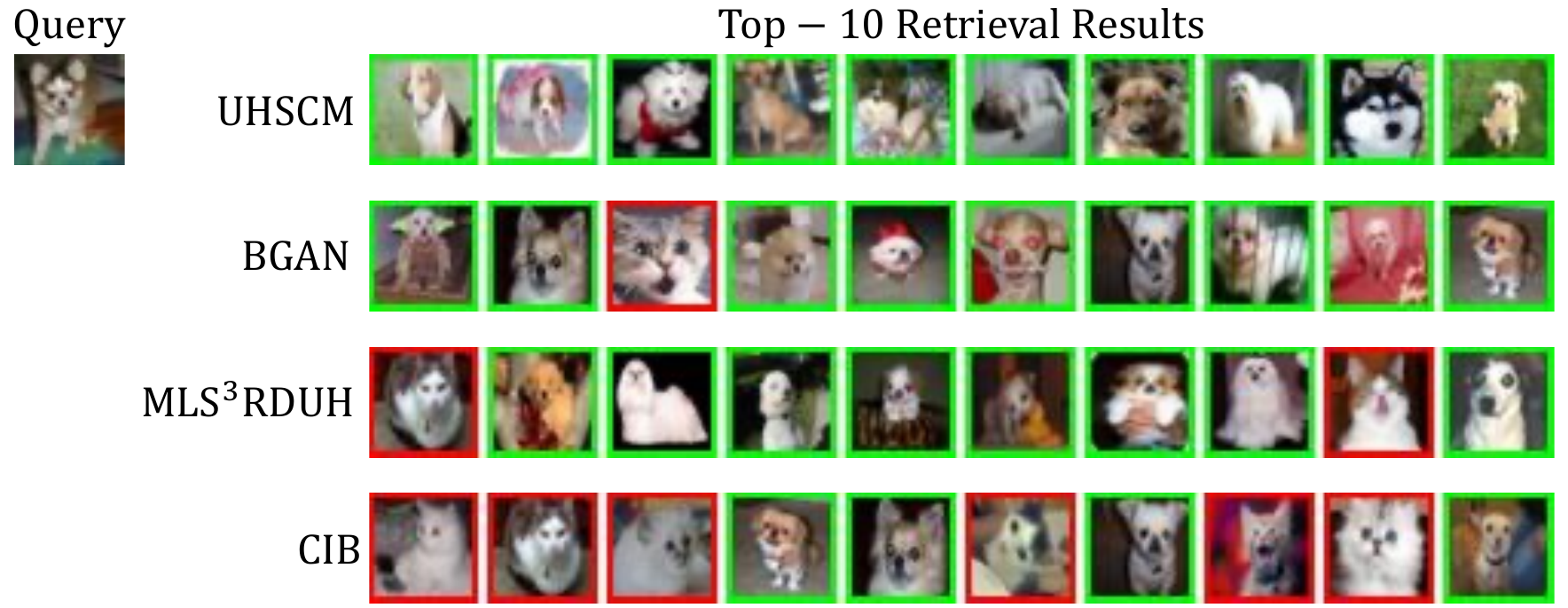}
	\caption{Top-10 retrieved results of UHSCM,  CIB, BGAN and MLS$^3$RDUH on the CIFAR-10 dataset.}
	\label{fig_case}
\end{figure}
\subsection{Time Consumption}

Here, we investigate the time consumption of our proposed method, and the results are shown in Table \ref{time}. In this experiment settings, for a method, its time consumption is the sum of the time spent on all its preprocessing operations and the time spent on training its hashing model to convergence. It can be found that the time consumptions of our proposed method UHSCM are 27.3, 35.7 and 20.2 minutes on the CIFAR10, NUS-WIDE and MIRFlickr-25K datasets, respectively, which are comparable to those of existing methods and  even much lower than the ones of BGAN and  MLS$^3$RDUH.

	\subsection{Sensitivity to Hyper-parameters}
	\label{shp}
	Here, we investigate the influence of hyper-parameters $\tau, \alpha, \lambda, \gamma$, and $\beta$  with the hash code length being 64 bits on the three datasets. 
	
	First, we study the influence of hyper-parameters $\tau$ over the three dataset with its value changed from 1$m$ to 4$m$ and the other hyper-parameters fixed, where $m$ denotes the number of concepts. The MAP results are shown in Figure \ref{fig_para} (a), (f) and (k). It can be found that our proposed method can achieves great retrieval performance on all the three datasets with $\tau$ being 1$m$ or 3$m$. Hence, in the other experiments, we set $\tau$ as 3$m$.
	
	Then, to study the influence of hyper-parameter $\alpha$, we vary  the  value of $\alpha$ from $0.1$ to $0.5$ with the other hyper-parameters fixed. As the results shown in Figure \ref{fig_para} (b), (g) and (l) for all the three datasets, UHSCM achieves good performance when $\alpha \in [0.1, 0.4]$. Moreover, for the CIFAR10 dataset, the proposed UHSCM obtains the best retrieval result when $\alpha=0.2$; for the NUS-WIDE dataset, the performance of UHSCM is the best when $\alpha= 0.1$; for the MIRFlickr-25K dataset, when $\alpha=0.3$, UHSCM achieves the best retrieval performance. 
	
	Next, we investigate the influence of hyper-parameter $\lambda$ over the three dataset with its value changed from 0.5 to 1.0 and the other hyper-parameters fixed, and the corresponding experimental results shown in Figure \ref{fig_para} (c), (h) and (m). It can be found  that when $ \lambda=0.8$, 0.5  and 0.6 our proposed method UHSCM can achieve the best performance for the CIFAR10, NUS-WIDE and MIRFlickr-25K datasets, respectively.
	
	Moreover, to study the hyper-parameter $\gamma$ on the three datasets, we vary the value of $\gamma$ from 0.1 to 0.6 with the other hyper-parameters fixed, and the results are shown in Figure \ref{fig_para} (c), (g) and (k). It can be seen that when $ \gamma=0.2$, our proposed method UHSCM can achieve the best performance for the CIFAR10 and NUS-WIDE datasets, and for the dataset MIRFlickr-25K, UHSCM obtains the  best retrieval performance when $\gamma=0.5$.
	
	Furthermore, we investigate the effect of hyper-parameter $\beta$ with its value varied from 0 to 0.1 and the other hyper-parameters fixed. The MAP results are shown in Figure \ref{fig_para} (d), (h) and (l). For all the three datasets when the $beta=0.001$, UHSCM achieves the best performance.
	
	Finally, based on the above experimental results, for the CIFAR10 dataset, we set $\alpha =0.2,\ \lambda=0.8,\ \gamma=0.2$, and $\beta =0.001$; for the NUS-WIDE dataset, the hyper-parameters $\alpha, \lambda, \gamma$, and $\beta$ are set as 0.1, 0.5, 0.2 and 0.001, respectively; for the MIRFlickr-25K dataset, we set hyper-parameters $\alpha, \lambda, \gamma$, and $\beta$ as 0.3, 0.6, 0.5 and 0.001, respectively.
	\subsection{Qualitative Results}
	\subsubsection{t-SNE visualization}
	To better understand the manifold structure of learned hashing code, we compare the t-SNE visualization \cite{van2008visualizing} of hash codes generated by our proposed UHSCM, CIB, MLS$^3$RDUH, and BGAN for the datapoints in the database of CIFAR10 dataset, and the results are shown in Figure \ref{fig_tsne} where the data points within the same colour belong to the same class. It can be easily found that compared with the three baselines,  our proposed UHSCM shows a clearer structure, in which the clusters of each class are separated from each other. There results demonstrate that compared with the baselines, our proposed method UHSCM can generate hash codes with more abundant semantic similarity information preserved, i.e., the generated hash codes are more distinguished.
	\subsubsection{Visualization of retrieval results}
	In Figure \ref{fig_case}, we show the top-10 retrieval results  of the proposed UHSCM, CIB, MLS$^3$RDUH and BGAN on the CIFAR-10 dataset with the length of hash codes being 64 bits. Specifically, the relevant results are framed by the green border, and the irrelevant results are framed by the red border. It can  be seen that comparing to these baselines, our DSAH has fewer fault images. These results show that the quality of hash codes generated by our proposed UHSCM are higher.
	
	\section{Conclusion}
	In this paper, we proposed a novel Unsupervised Hashing with Semantic Concept Mining, dubbed  UHSCM.  UHSCM first leverages the CLIP model to denoise a set of randomly collected concepts according to the available training images through the prompt engineering. Next, based on the denoised concepts, UHSCM mines the concept distribution of each image by the VLP model through prompting again to constructs a high-quality semantic similarity matrix. Finally, treating the constructed semantic similarity matrix as guiding information, a novel hashing loss with a modified contrastive loss based regularization item was proposed to optimize the hashing network. Extensive experiments on three benchmark datasets have shown that the proposed method outperforms the state-of-the-art baselines on the image retrieval task.
	
	\section*{Acknowledgments}
	The work is supported by National Key R\&D Plan (No. 2020AAA0106\\600), National Natural Science Foundation of China (No. U21B2009, 62172039 and L1924068).
	\bibliographystyle{ACM-Reference-Format}
	\bibliography{acmart}
\end{document}